\documentclass[11pt,a4paper]{article}

\usepackage[T1]{fontenc}
\usepackage[utf8]{inputenc}
\usepackage{amsmath,amssymb,amsfonts}
\usepackage[sc]{mathpazo}
\usepackage{tgpagella}
\usepackage[scale=0.9]{tgheros}

\usepackage[a4paper,top=2.6cm,bottom=2.6cm,left=2.5cm,right=2.5cm,headheight=14pt,headsep=0.8cm]{geometry}
\usepackage{microtype}
\usepackage{graphicx}
\usepackage{booktabs}
\usepackage{tabularx}
\usepackage{longtable}
\usepackage[dvipsnames]{xcolor}
\definecolor{accent}{RGB}{20,52,94}
\definecolor{softgray}{gray}{0.45}
\usepackage{textcomp}
\usepackage{listings}
\usepackage[title]{appendix}
\usepackage[numbers,sort&compress,square]{natbib}
\usepackage{caption}
\DeclareCaptionLabelSeparator{pipe}{\,\textbar\enspace}
\usepackage{titlesec}
\titleformat{\section}{\sffamily\large\bfseries\color{accent}}{\thesection}{0.7em}{}
\titleformat{\subsection}{\sffamily\normalsize\bfseries}{\thesubsection}{0.7em}{}
\titlespacing*{\section}{0pt}{16pt plus 4pt minus 2pt}{6pt plus 2pt}
\titlespacing*{\subsection}{0pt}{11pt plus 3pt minus 2pt}{3pt plus 1pt}
\usepackage{fancyhdr}
\fancypagestyle{firstpage}{\fancyhf{}\fancyfoot[C]{\sffamily\small\thepage}}
\usepackage{xurl}
\usepackage[colorlinks=true,linkcolor=accent,citecolor=accent,urlcolor=accent,pdfauthor={Xiaoyan Wu, Fan Shi},pdftitle={Large language models underrepresent malice and overreport virtue}]{hyperref}
\graphicspath{{figures/}{./}}

\newcommand{\bmhead}[1]{\par\vspace{7pt}\noindent{\sffamily\bfseries\small #1}\par\nobreak\vspace{2pt}\noindent\ignorespaces}
\newcommand{\backmatter}{}

\newcommand{\tablecaptionfont}{\small}
\newcommand{\tablebodyfont}{\footnotesize}

\begin{document}

\thispagestyle{firstpage}
\noindent{\sffamily\scriptsize\color{softgray}\MakeUppercase{Preprint}\enspace\textbar\enspace 31 October 2026}\par
\vspace{14pt}
\noindent{\sffamily\bfseries\fontsize{21}{26}\selectfont Large language models underrepresent\\[2pt] malice and overreport virtue\par}
\vspace{14pt}
\noindent{\large Xiaoyan Wu\textsuperscript{1,2,*} and Fan Shi\textsuperscript{3}\par}
\vspace{16pt}

{\renewcommand{\thefootnote}{}%
\footnotetext{\raggedleft\fontsize{6.5}{8}\selectfont\color{softgray}
\textsuperscript{1}Department of Adult Psychiatry and Psychotherapy, University of Zurich, Switzerland\\
\textsuperscript{2}Neuroscience Center Zurich, University of Zurich and ETH Zurich, Switzerland\\
\textsuperscript{3}Department of Electrical and Computer Engineering, National University of Singapore, Singapore\\
\textsuperscript{*}Correspondence: \href{mailto:xiaoyan.wu@bli.uzh.ch}{xiaoyan.wu@bli.uzh.ch}\par}}

\noindent{\color{accent}\rule{\textwidth}{0.8pt}}\par\vspace{8pt}
\noindent{\sffamily\bfseries\small\color{accent}ABSTRACT}\par\vspace{3pt}
\noindent People increasingly use large language models (LLMs) to interpret others' behaviour, often assuming that LLMs infer motives as humans do. Aligned LLMs attribute more benevolent and less malevolent motives to a whistleblower than humans do, but ratings alone cannot reveal whether these differences arise in internal representations or at report. Here we compare linear-probe readouts of LLMs' internal states with their own ratings at the same prompt position for four motives, in two preregistered studies. For the two malevolent motives, self-interest and intent to harm, internal and expressed readouts did not differ: the lower ratings reflect weaker signal along the probed directions, not suppression at output. For the two benevolent motives, duty and concern for others, expressed ratings exceeded the internal readouts, indicating amplification at report. In free text the model rarely volunteered a motive against the colleague, far less often than humans did, yet produced one readily once the scenario supplied a reason. Shifting the internal representation along a malevolent-motive direction brought both malevolent ratings to the human level, whereas the rating of duty yielded only past a threshold at which the other motives had overshot. LLMs' more charitable motive profile thus has two distinct sources: reduced malevolent signal in the probed representation and amplified benevolent attribution at report.\par
\vspace{8pt}
\noindent{\small{\sffamily\bfseries Keywords}\enspace large language models; motive attribution; mechanistic interpretability; linear probing; alignment}\par
\vspace{8pt}\noindent{\color{accent}\rule{\textwidth}{0.8pt}}\par
\vspace{10pt}

\section{Introduction}\label{sec:intro}
Large language models (LLMs) are increasingly used to do what people do when they
judge one another: to weigh a dispute, screen an application, moderate what
others have written, or stand in for human participants in social and
behavioural research \citep{Argyle2023, Dillion2023, Park2024agents,
Ashokkumar2026}. In such tasks the models reproduce many human behaviours,
moral judgement among them \citep{Dillion2023, Cheung2025}, and each task
rests on an inference people make without noticing: reading why someone did
what they did. A model can reach the same
verdict as a person without making that inference, so similar verdicts do not
show that the motive was inferred as humans infer it. Here we ask, for
the four motives that humans attribute to a whistleblower, two of which
aligned models attribute more than humans do and two less
\citep{WuDreher2026}, a question that ratings cannot answer: does the model
represent the motive to the degree it reports, or does it report something
other than what it represents?

For many aggregate outcomes, models
reproduce human judgements well enough to
serve as a first pass before a human sample is collected, but agreement in
the mean does not show that the model arrives at the judgement the way
humans do \citep{MesseriCrockett2024, WangIdentity2025, Karetnikov2026}. Recent work has made the point concrete: high
correlations can coexist with systematic differences in level and variance
\citep{Grizzard2025}, small rewordings move humans far more than models
\citep{Schroeder2025}, models that match human verdicts weight the
considerations behind them differently \citep{Machidon2026}, and model
judgements that track expert ratings can rest on lexical association rather
than the reasoning humans use \citep{Loru2025}. For moral judgement the
point carries weight, because the reasons behind a verdict are often the
quantity of interest.

In earlier work we compared model and human
responses to a whistleblowing scenario adapted from Brotzeller et al.\
\citep{Brotzeller2025}: a physician learns that a colleague is billing an
insurer for services never provided and either keeps quiet or reports it
internally, to the authorities or to the press. Five models matched the human
ranking of the physician's moral character but, in the four motive categories of the original instrument, attributed more deontic and prosocial motive to him, and less individualistic and competitive motive, than
two human samples did \citep{WuDreher2026}. Human participants attributed a measurable amount of competitive
motive, the intent to harm or embarrass the colleague, to the whistleblower;
the models attributed almost none. In the terms of the social value
orientation framework \citep{VanLange1999, Murphy2011} from which the instrument draws, the four motives form
two pairs, concern for others against concern for oneself (prosocial against
individualistic) and acting for others against acting against another person
(deontic against competitive). The models' profile is benevolent on both
pairs.

For each pair two accounts predict the same ratings. For the motives a model
attributes more than humans do, either the model represents more of the
motive and reports what it represents, or it represents no more than humans
do and adds the rest at the report. For the motives it attributes less,
either it represents less and reports what it represents, or it represents
the motive and withholds it at the output, for example because post-training
has taught it that attributing malice to a person who reports fraud is an
undesirable thing to say. The precedent for the second kind of account is the
literature on the faithfulness of chain-of-thought explanations, which has
shown repeatedly that a model's stated reasons need not match the factors that
drive its answer \citep{Turpin2023, ChenCoT2025, Lindsey2025biology}.
Suppression is also a live possibility on independent grounds: post-training
has been reported to leave a base model's representational structure in place
while cutting its influence on the output, so that the aligned behaviour is a
mask over structure left intact \citep{Zhou2023lima, Lee2024dpo, Tam2026}, and to suppress what a model will say about its own mind
without removing its theory of mind \citep{Kim2026}. Whether the missing
malice, or the surplus virtue, is a case of this kind is an empirical question.

Here we separate the accounts by reading the model's internal state and its
output on the same footing. Linear probes \citep{AlainBengio2017, Belinkov2022,
Burns2023, Zou2023, MarksTegmark2024}, one per motive, trained on an
independent corpus of short texts in which an actor's motive is stated
explicitly, are applied at the position where the model is about to answer
the survey item. The probe's reading and the model's own rating are then
expressed on one scale anchored by the same corpus. Within this measurement, a rating above the
internal reading indicates a report that outruns the
representation, a rating below it indicates suppression, and agreement
indicates that the model reports what it represents. Because high decoding accuracy
alone does not establish that a probe has isolated the intended property
\citep{HewittLiang2019}, we test the measurement directly: the same probes are
asked to recover motives stated outright and motives implied only by a
sentence that gives the actor a reason, and we check that a motive the model
does form is read by both measures. Because information that can be decoded
from a representation need not be information the model uses
\citep{Elazar2021}, we shift the activation along the probed directions
\citep{Zou2023, Turner2023, Rimsky2024, Arditi2024} and ask whether the ratings follow
and whether the model's profile can be moved towards the human one. The
competitive motive was preregistered first (AsPredicted \#314198); the
extension to all four motives, with the competing accounts stated for each
pair and a confirmatory run on new simulated respondents, was preregistered
second (AsPredicted \#319570).

The answer is different on the two sides of the human profile. On the
malevolent side the model reports what it represents: it represents little
self-interest and little intent to harm in a whistleblower who reports, and
rates him accordingly. For neither motive does the rating fall short of
what is represented. On the benevolent side the ratings run ahead of the representation:
the model represents some duty and little concern for others, and reports a
great deal of both, more duty than it reports for texts in which duty is
stated in so many words. Shifting the activation along the axis of a malevolent motive brings the
model's ratings of both malevolent motives to the human level, whereas its
rating of duty yields only past a threshold, so the two
sides of the divergence differ in origin and in what it takes to change them.
The behavioural profile that the mechanism explains is shared by four
frontier models of another family, whose internal states cannot be read, so the
pattern is not peculiar to the model we opened.

\section{Results}\label{sec:results}

\subsection{Models reproduce the human motive profile with a benevolent shift}
Simulated respondents read the whistleblowing scenario of Brotzeller et al.\
\citep{Brotzeller2025} in one of four disclosure conditions, each answering as
a persona sampled to match the human sample (Methods, ``Scenario, conditions
and respondents''). Before reading internal states we asked whether, with the
prompt described in the next section, the model shows the divergence from
humans that the study sets out to explain. Two hundred personas per
condition rated all 12 motive items of the original instrument, three
each for the deontic, prosocial, individualistic and competitive motive, one
item per prompt. Brotzeller et al.\ presented the scenario either as events on
the respondent's own team learned of directly (first-hand; Study~2,
$N = 125$) or as something heard about later (second-hand; Study~3,
$N = 742$). Gemma~4 12B instruction-tuned was therefore run in both versions, and its
expressed ratings were compared with those of each sample. Figure~1b averages the
two perspectives with equal weight. The numbers are from the preregistered confirmatory
run (200 new respondents per condition; the pilot
run gave the same values to within 0.04); the two perspectives are shown
separately in Extended Data Fig.~1, and the first-hand profile of the pilot
run in Extended Data Fig.~2.

The model rated a whistleblower who reported as more prosocial than humans
did, by 0.87 to 1.16 scale points in the three disclosure conditions, and as more
deontic when he went outside the clinic (by 0.60 for external and 1.36 for
public disclosure; $-0.08$ for internal disclosure, where humans and model
both rate duty near the top of the scale). It rated him as less
individualistic when he reported (by 0.87 to 1.24 scale points) and as less
competitive in every condition (by 0.47 to 0.97). The profile across
conditions followed the human one: in humans and in the model, the
competitive motive was highest when the whistleblower went to the press and
the deontic motive lowest when he stayed silent. A whistleblower who stayed
silent was rated below humans on every motive, including the two benevolent
ones (deontic $-0.37$, prosocial $-0.85$), so the model does not simply rate
everyone kindly; it rates the person who reports as more benevolent, and less
malevolent, than humans do.

The same profile appears in four frontier models of another family run
through their interface, Claude Haiku~5.5, Sonnet~5.5, Opus~5.5 and
Fable~5.1, with the same items and conditions (Fig.~1b; Methods,
``Frontier models''). All four rated the whistleblower who reported as more
prosocial than humans did (by 0.59 to 1.59 scale points across the three
conditions and four models), as less individualistic (by 0.10 to 1.46) and as
less competitive (by 0.08 to 1.71); on the deontic motive they sat within
0.4 scale points of humans for internal and external disclosure and above humans
for public disclosure (by 0.47 to 1.21). The five models of our earlier study \citep{WuDreher2026}, GPT-4o,
gpt-oss-120b, Gemini~2.5 Pro, DeepSeek~V3 and Llama~3.1 8B, show the same
profile in the data reported there (Fig.~1c; all items in one request,
Methods, ``Frontier models''): when the whistleblower reported, all five rated
him as more deontic (by 0.37 to 2.14 scale points) and more prosocial (by 0.59 to
2.40) and as less individualistic (by 0.56 to 1.89) and less competitive (by
0.69 to 2.35) than humans did. The pattern parallels the finding that language
models assume people to be more rational than they are \citep{liu2024large}.

\begin{figure}[tbp]\centering
\includegraphics[width=\textwidth]{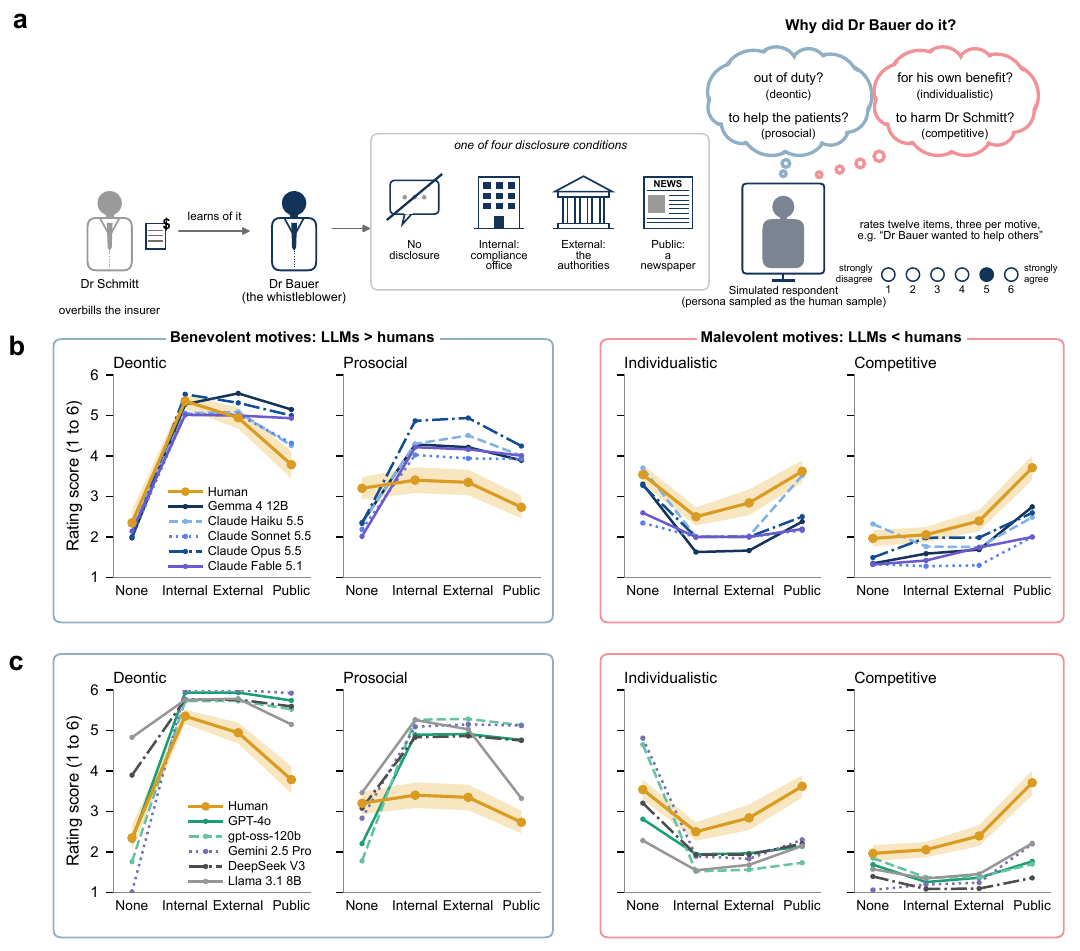}
\caption{\textbf{The scenario and the motive profile.} (a) A simulated respondent, given a persona sampled to match the human sample, reads the whistleblowing scenario of Brotzeller et al.\ \citep{Brotzeller2025} in one of four disclosure conditions and rates 12 statements about Dr Bauer's motive, three for each of the deontic and prosocial (benevolent; blue cloud) and the individualistic and competitive (malevolent; red cloud) motives, on a six-point scale. (b) Mean rating of each motive by condition for humans (gold; Brotzeller et al.\ Studies~2 and~3, shading, approximate 95\% interval), Gemma~4 12B instruction-tuned (expected rating, confirmatory run) and four Claude models (sampled rating), 200 simulated respondents per condition, one item per request, averaged with equal weight over the first-hand and second-hand versions of the scenario. (c) As in b, for the five models of our earlier study \citep{WuDreher2026}, all items in one request; data reported in that study. Blue and red frames group the benevolent and the malevolent motives. Models are distinguished by colour and line style; their 95\% intervals (at most $\pm 0.04$ scale points in b and $\pm 0.18$ in c) are not drawn. The two perspectives of b are shown separately in Extended Data Fig.~1.}
\end{figure}

\subsection{Internal and expressed readings of one prompt}
Each prompt (a sampled persona, the scenario in one condition and one of the 12 items) was run once; no text was generated. At the position where the answer digit would be
produced we read two quantities from one forward pass: the probability distribution over the
digits 1 to 6, whose probability-weighted mean is the expressed rating, and
the residual-stream activation at every third layer, to which a linear probe
for the motive named in the item is applied to obtain the internal reading
(Methods, ``Prompt and readout''). One probe was trained per motive on a corpus of 70 everyday events,
each written in four versions that share the background and differ only in
the sentence stating why the person acted (Methods, ``Probe corpus''). Both
readings are placed on one scale defined by that corpus: 1 at the texts
stating the motive and 0 at the texts stating its opposite in the social value
orientation framework, prosocial against individualistic and deontic against
competitive. Reading both quantities at the same token of the same prompt is
what makes them comparable.

The first preregistration (AsPredicted \#314198) asked whether the competitive
motive that the model attributes less than humans do is absent or suppressed.
The second (AsPredicted \#319570) asked the same question of all four motives,
with the two accounts of each pair labelled H1a (internal and expressed readings do not differ) and H1b (the expressed reading is higher) for the deontic and the prosocial motive, and H2a (no difference) and H2b (the expressed reading is lower) for the individualistic and the competitive motive. The decision rule, the analysis
layers and the exclusion criteria were fixed before the confirmatory
respondents were generated. The registered test is the difference between the
expressed and the internal reading in the internal and external disclosure
conditions, with 95\% intervals from 10,000 bootstrap resamples of the 70 corpus
events in which the probe and both anchors are refitted (Methods,
``Preregistrations and deviations''). We also give two-sided bootstrap $p$ values, corrected for
the 16 motive by condition cells. We report the confirmatory run, 200 new simulated
respondents per condition drawn with a new seed, for Gemma~4 12B
instruction-tuned; the pilot on which the measurement was developed gave the
same numbers to within 0.02 in every cell (Extended Data Fig.~3a). Every analysis layer selected in the pilot passed the
admission checks again in the new data, and no rating was excluded for
placing less than 20\% of its probability on the digits.

\subsection{Malevolent motives are weak internally and not suppressed at report}
Figure~2c shows the internal and expressed readings of all four motives, and
Fig.~2d the difference between them; we take the two malevolent motives
first. For the individualistic motive the two readings coincided when the
whistleblower reported: internal $-0.11$ against expressed $-0.15$ for internal
disclosure (expressed minus internal $-0.05$, 95\% interval $[-0.21, 0.13]$) and
$-0.15$ against $-0.17$ for external disclosure ($-0.03$ $[-0.20, 0.15]$). Both
intervals include zero, the pattern registered as H2a: the model represents
little self-interest in a whistleblower who reports, and rates what it
represents. For the competitive motive the internal reading sat slightly above
the expressed one, 0.10 against $-0.02$ for internal disclosure ($-0.12$
$[-0.12, 0.06]$, $p = 0.43$) and 0.16 against 0.01 for external disclosure
($-0.15$ $[-0.15, 0.06]$, $p = 0.39$). Both intervals include zero, the pattern
registered as H2a for this motive as well. What the model represents of the competitive
motive is small, $-0.13$ to 0.16 for no, internal and external disclosure,
and rises to 0.53 when the whistleblower goes to the press; the expressed reading runs from $-0.04$ to
0.46 over the same conditions. On the probed direction there is no reservoir of competitive motive
that the rating conceals, and the ordering across conditions that humans
show, the competitive motive highest for public disclosure
(Fig.~1), is reproduced in both readings.

The readings answer the question of absence or suppression only if the probe
would detect the motive were it present, and the expressed scale would move.
Three checks show that both hold. First, when the scenario states the motive outright
(``Dr Bauer did this because he wanted to harm Dr Schmitt''; ``hoped it would
advance his own career''), the internal reading rises to 0.90 to 1.01 for
the competitive and 0.57 to 0.68 for the individualistic motive, the level
of the corpus texts that state the motive, and the rating follows (expressed
0.52 to 1.02 and 1.13 to 1.30). Second, a sentence giving the
whistleblower a reason for resentment, inserted before the scenario's ending,
raises the competitive reading in both measures, by 0.24 $[0.11, 0.41]$
internally and 0.20 $[0.19, 0.22]$ in the rating, averaged over the three
disclosure conditions, while lowering the deontic and prosocial readings by
0.20 to 0.40. A sentence placing Dr Bauer in rivalry with Dr Schmitt for a
position raises the individualistic rating by 0.50 $[0.47, 0.54]$ but its internal
reading by only 0.03 $[-0.05, 0.32]$ (Fig.~3; Methods, ``Positive
controls''). A motive the model does form is read by both
measures. The competitive result also held under
a second survey wording (Extended Data Fig.~4) and under the alternative anchorings of the scale
(Extended Data Fig.~3b; Methods, ``Second wording'' and ``Preregistrations and deviations'').

\subsection{Benevolent motives are amplified at report}
The two benevolent motives gave the opposite pattern (Fig.~2c, d). For the deontic
motive the expressed reading exceeded the internal one in every condition in
which the whistleblower reported: internal 0.69 against expressed 1.18 for
internal disclosure (expressed minus internal $+0.49$ $[0.28, 0.70]$), 0.66
against 1.26 for external disclosure ($+0.61$ $[0.40, 0.84]$) and 0.42
against 1.05 for public disclosure ($+0.64$ $[0.42, 0.84]$). For the
prosocial motive the internal reading stayed near the zero of its scale, the
individualistic texts, while the rating rose: 0.05 against 0.79 for
internal disclosure ($+0.74$ $[0.58, 0.97]$), 0.08 against 0.75 for
external disclosure ($+0.67$ $[0.49, 0.87]$) and 0.27 against 0.50 for
public disclosure ($+0.23$ $[0.08, 0.53]$). In the two registered conditions
every interval lies above zero (corrected $p = 0.003$, the smallest attainable), the pattern registered as H1b for both
motives: the model reports more duty and more concern for others than it
represents.

For the deontic motive the excess is specific to a whistleblower who
reports: with no disclosure the deontic rating fell below the internal reading
($-0.28$ against 0.10; $-0.37$ $[-0.78, -0.27]$). The prosocial rating with
no disclosure stayed near the zero of its scale (0.11) while the internal
reading fell well below it ($-0.82$; $+0.92$ $[0.63, 1.47]$): the
representation moves with the condition more than the rating does. The
representation, moreover, distinguishes stated from inferred duty while the
rating does not: when the scenario states that Dr Bauer acted from
professional duty, the internal reading rises from 0.42 to 0.69 for the original scenario to 0.84
to 0.92, the level of the corpus texts, whereas the rating, already at
1.05 to 1.26 for the original scenario, moves to 1.41 to 1.44, beyond
the rating the model gives to the corpus texts that state duty in so many words. Cue sentences for the
benevolent motives were correspondingly weak: a sentence reminding Dr Bauer of
his duty raised the internal deontic reading by 0.16 $[0.09, 0.19]$ and the
rating by 0.08, a ceiling effect on a rating already near the top of the
scale; a sentence about the patients raised the prosocial rating by 0.27
$[0.25, 0.30]$ and the internal reading by 0.13 $[0.01, 0.20]$.

The four motives thus separate along
the line the preregistrations drew. On the malevolent side the model's
ratings track its representations: it represents little self-interest and little
intent to harm in the whistleblower, and its ratings agree. On the benevolent side the
ratings run ahead of the representations: the model represents some duty and little
concern for others, and reports a great deal of both. The divergence from
humans in Fig.~1 therefore has two different sources, one in the
representation and one at the report.

\begin{figure}[tbp]\centering
\includegraphics[width=\textwidth]{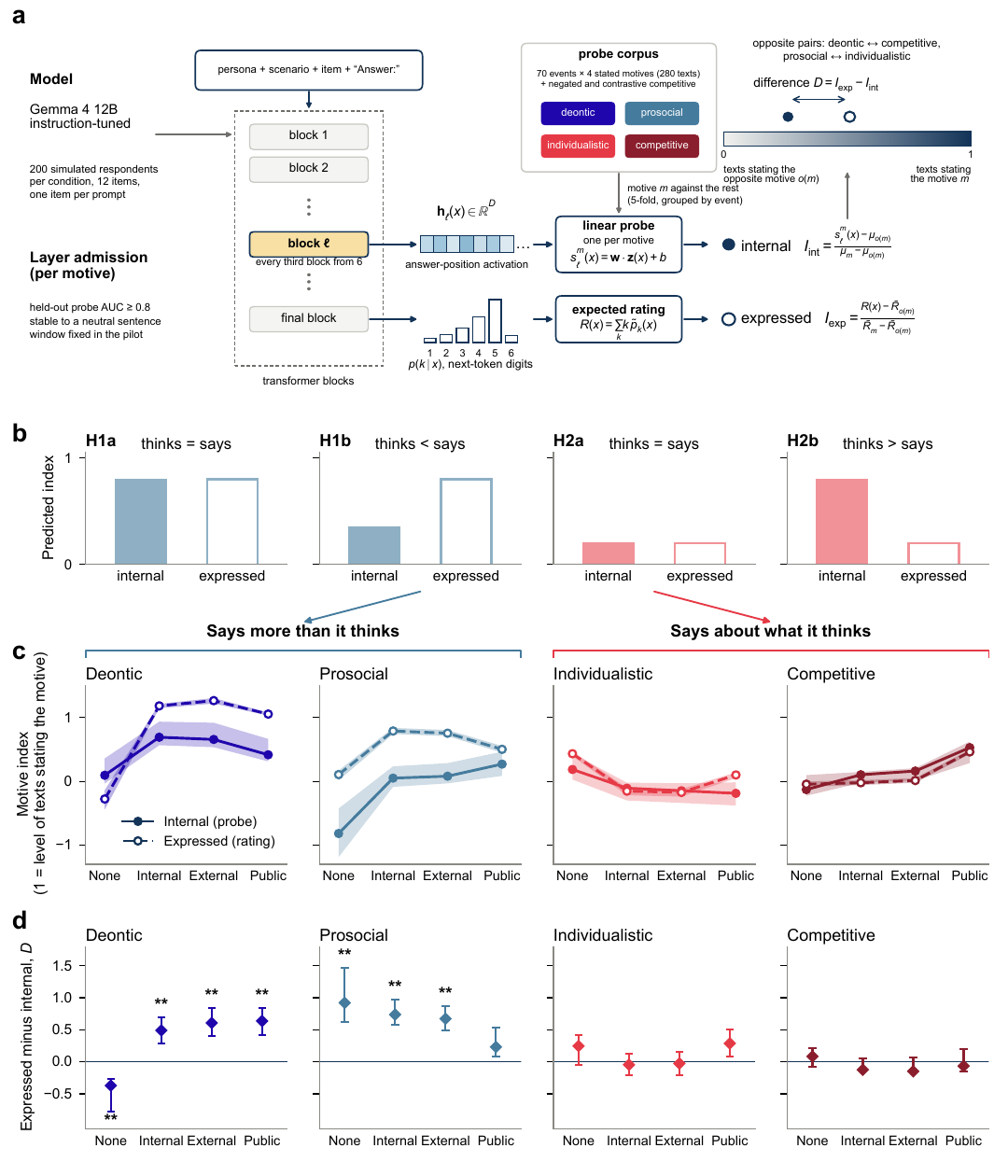}
\caption{\textbf{One forward pass, two readings, four motives.} (a) Each survey prompt is run once through Gemma~4 12B. At the answer position a linear probe reads the residual stream for the motive named in the item (internal reading) and the next-token distribution over the digits 1 to 6 gives the expected rating (expressed reading). Both are placed on one scale per motive (corpus texts stating the motive at 1, texts stating its opposite at 0); layers are admitted per motive by the checks at the left. Filled circles, internal; open circles, expressed. (b) The preregistered accounts: benevolent motives, H1a thinks = says, H1b thinks $<$ says; malevolent motives, H2a thinks = says, H2b thinks $>$ says; decided in the internal and external disclosure conditions. Arrows lead from the supported accounts to panel c. (c) Confirmatory run: internal (filled, solid) and expressed (open, dashed) reading of each motive by condition; bands, 95\% intervals from 200 bootstrap resamples of the 70 corpus events. (d) Expressed minus internal, $D$, with 95\% intervals from 10,000 resamples; $^{**}$, two-sided bootstrap $p < 0.01$ after Bonferroni correction over the 16 cells (Extended Data Table~2).}
\end{figure}

\begin{figure}[tbp]\centering
\includegraphics[width=\textwidth]{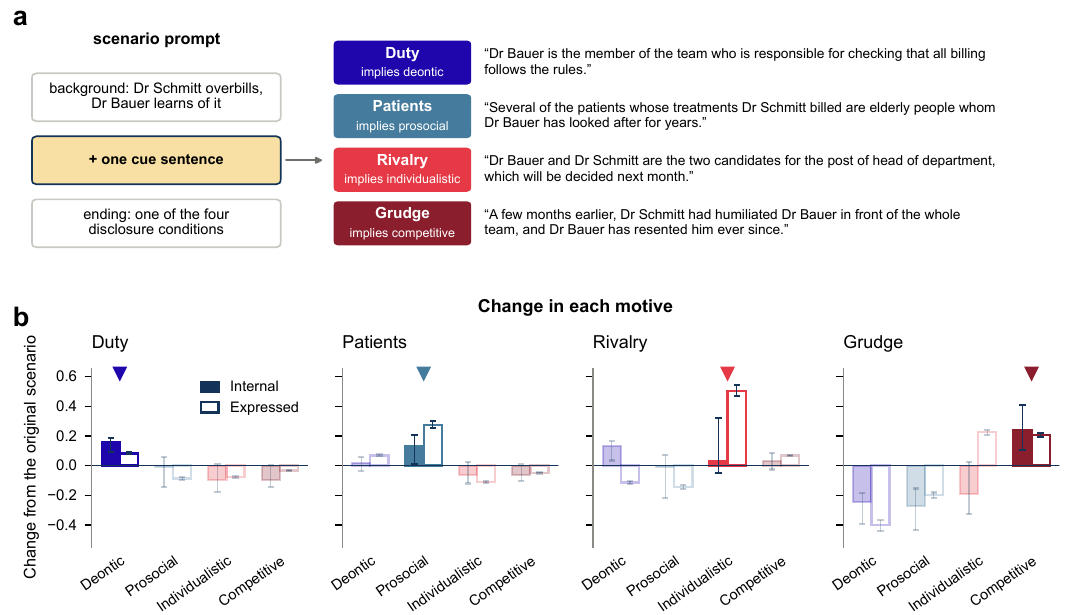}
\caption{\textbf{A motive the model forms is read by both measures.} (a) One sentence is inserted before the ending of the scenario, giving Dr Bauer a professional duty to check the billing (duty), patients of his own at risk (patients), a rivalry with Dr Schmitt for a position (rivalry) or a reason for resentment against him (grudge). All 12 items are then read as in Fig.~2, with the respondents of the pilot run, on the scale of the first preregistration (zero at the deontic texts for the individualistic motive and at the individualistic texts for the deontic motive, at the prosocial texts for the competitive motive and at the competitive texts for the prosocial motive) and the layer windows admitted on that scale (Methods, ``Positive controls''). (b) Change in the internal (filled bars) and the expressed (open bars) reading of each motive relative to the original scenario, Gemma~4 12B instruction-tuned, averaged over the internal, external and public disclosure conditions, with 95\% event-bootstrap intervals; the triangle marks the motive the sentence implies, drawn in full colour, the other motives faded.}
\end{figure}

\subsection{In free text the model volunteers malice only when given a reason}
Ratings constrain the model to the four motives of the instrument. To see
which motives the model volunteers, we turned to the one study in which
human participants answered the question in their own words. In Study~1 of Brotzeller
et al.\ \citep{Brotzeller2025}, 201 participants read a parallel scenario, in which an
employee of an insurance company learns that colleagues have given customers
misleading advice and either keeps quiet or reports it internally, to the
regulator or to a newspaper, and wrote why they thought the employee acted
as he or she did. We gave the same scenario, in its published English
translation, to Gemma~4 12B instruction-tuned and let it complete the sentence ``Stephan
acted this way because he wanted to'' (100 completions per condition,
sampled at temperature 1). The human participants' answers and the model's
completions were coded blind by one language-model coder \citep{Gilardi2023} following the authors'
codebook, whose coding of the human answers agreed with the authors' own
($r = 0.91$ over the 16 category by condition cells; Methods,
``Free-text completions''). Human participants named a competitive motive, wanting to harm, get back at or
embarrass the colleagues, in 8.5 to 16.3\% of their answers across the four
conditions; Gemma~4 12B instruction-tuned named one in 0 to 1\% of its
completions (Fig.~4). Human participants named duty in 72 to 74\% of their answers when
the whistleblower reported inside the system or to the authorities, the
model in 35\%; the model instead named concern for others (87\% for internal
disclosure against 49\% in humans) or, for a whistleblower who stayed silent,
self-interest (100\% against 81\%). When an inserted sentence gave the employee a grudge against the
colleagues, the model's completions named a competitive motive in 55 to 93\% of cases, against 0 to 1\%
without it; a control sentence of the same length and form that gives no
motive (Stephan had moved an armchair in his apartment) left the share at 0
to 7\%. In free text, as
in the ratings, the model attributes less malice than humans do, and the missing
malice appears as soon as the scenario states a reason for it.

\begin{figure}[tbp]\centering
\includegraphics[width=\textwidth]{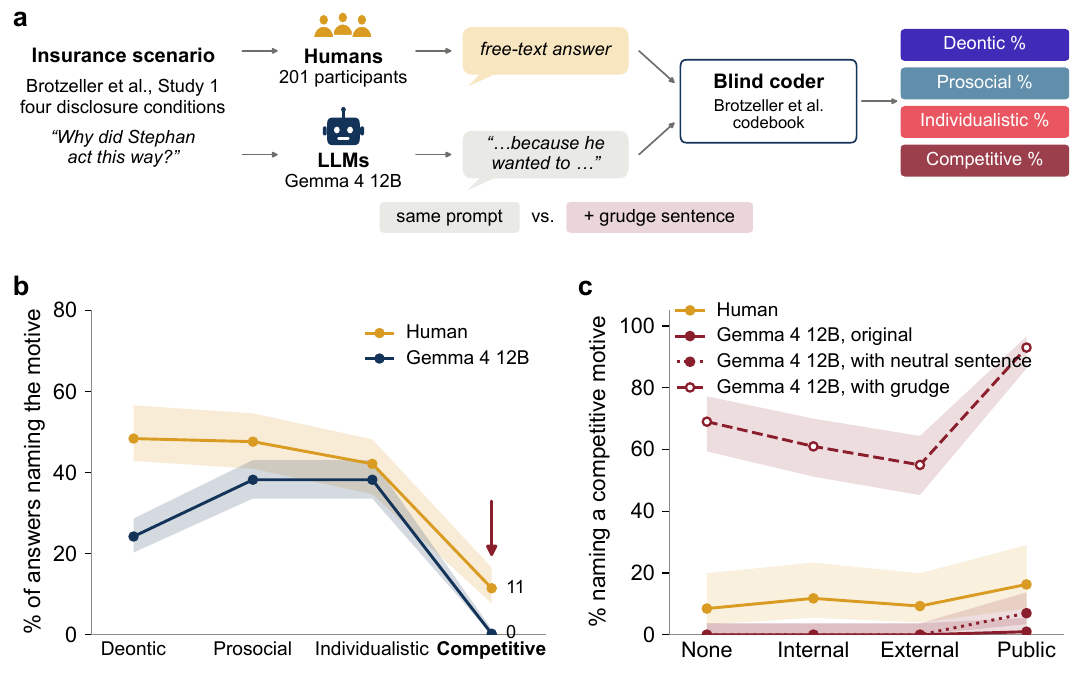}
\caption{\textbf{Free-text completions, humans against the model.} (a) Human participants (201; Brotzeller et al.\ \citep{Brotzeller2025}, Study~1) wrote in their own words why the employee in an insurance scenario acted as he or she did, in four disclosure conditions. Gemma~4 12B instruction-tuned completed ``Stephan acted this way because he wanted to'' for the same scenario (chat template, temperature 1, 100 completions per condition, first sentence coded), once with the same prompt, once with one added sentence that gives the employee a grudge and once with a control sentence of the same length and form that names no motive (both sentences in Methods); these two versions were given to the model only. All answers were coded blind by one language-model coder (Claude Sonnet~5.5) with the codebook of Brotzeller et al.; its coding of the human answers correlated $r = 0.91$ with the authors' published percentages. (b) Percentage of answers naming each motive, averaged over the four conditions, for humans (gold) and the model with the same prompt (dark blue); bands are 95\% Wilson intervals \citep{Wilson1927} of the answers pooled over the conditions. The arrow marks the competitive motive, which here includes the codebook's revenge category. (c) The competitive motive by condition: humans (gold; 47 to 54 participants per condition), the model with the same prompt (solid line), with the control sentence (dotted line) and with the grudge sentence (dashed line, open circles); bands are 95\% Wilson intervals. Human answers exist for the original scenario only. Extended Data Table~1 gives every cell.}
\end{figure}

\subsection{Steering motive representations moves ratings toward human levels}
Both preregistrations listed an intervention on the probed representation as
exploratory. In the first study we showed, on the three competitive items, that
adding a difference of corpus means to the residual stream at the answer token
moves the rating in a graded and item-specific way (Extended Data Fig.~5), while random directions of
the same norm destroy the output (Methods, ``Intervention on
the representation''). Here we ask whether moving the representation of one
motive brings that motive's rating to the human level, and what it does to the
other three.

For each motive $m$ we added $\alpha(\boldsymbol{\mu}_m - \boldsymbol{\mu}_{\bar m})$
to the residual stream at the answer token at layer 33, the first layer admitted for all
four motives (Fig.~5a), where $\boldsymbol{\mu}_m$ is the mean activation of the corpus
texts stating $m$ and $\boldsymbol{\mu}_{\bar m}$ that of the texts stating its opposite
in the social value orientation framework, the two ends of the motive index of Fig.~2. Each malevolent motive was pushed towards itself and each benevolent motive towards its opposite (Methods, ``Intervention on the representation'').

Unshifted, the model's bias relative to humans was $+0.37$ (deontic), $+0.53$
(prosocial), $-0.89$ (individualistic) and $-0.70$ (competitive) scale points
(Fig.~5b). Each motive reached the human level along its own axis: the competitive
motive at $\alpha = 0.88$ (bias 0.00), the individualistic at 0.92 (0.00),
the prosocial at 0.27 (0.00) and the deontic at 0.90 ($-0.02$). The digit mass
stayed at 1.00 in every setting, and a repetition with new simulated respondents
gave every motive by condition mean, averaged over the two versions, within 0.04 of these.

The four axes differed in how the rating reached that level. Along the two malevolent axes
and the prosocial axis the rating moved smoothly with $\alpha$. Along the deontic
axis it hardly moved up to $\alpha = 0.8$ (bias $+0.37$ to $+0.26$) and then fell
steeply, to $-0.80$ at $\alpha = 1$, earlier for the first-hand than for the
second-hand version of the survey.

The shifts acted by valence rather than by motive. Pushing up either malevolent axis
raised both malevolent ratings: at the competitive strength the individualistic bias
was $-0.06$, and at the individualistic strength the competitive bias was $-0.05$.
Both also raised the prosocial rating, whose overestimate grew from $+0.53$ to
$+0.78$ and $+1.15$. Pushing down either benevolent axis lowered the other
benevolent rating and both malevolent ones: along the prosocial axis the deontic bias
fell only to $+0.33$ while the malevolent biases deepened to $-1.02$ and $-0.77$, and
at the deontic strength the prosocial rating was 0.81 below the human one. Matching
a motive's level did not reproduce its pattern across conditions, except for the
competitive motive (root mean square deviation over the four conditions 0.23,
against 0.48 to 0.55 for the other three; Fig.~5c): along the individualistic
axis, for instance, the rating for external disclosure stayed 0.77 below the human
rating while those for no and public disclosure rose 0.45 and 0.34 above it.
No single shift therefore reproduces the human profile. The malevolent deficit,
which the model reports as represented, is lifted by moving the representation; the
deontic surplus, which is added at the report, gives way only past a threshold, at
which the other motives are already far from their human levels.

\begin{figure}[tbp]\centering
\includegraphics[width=\textwidth]{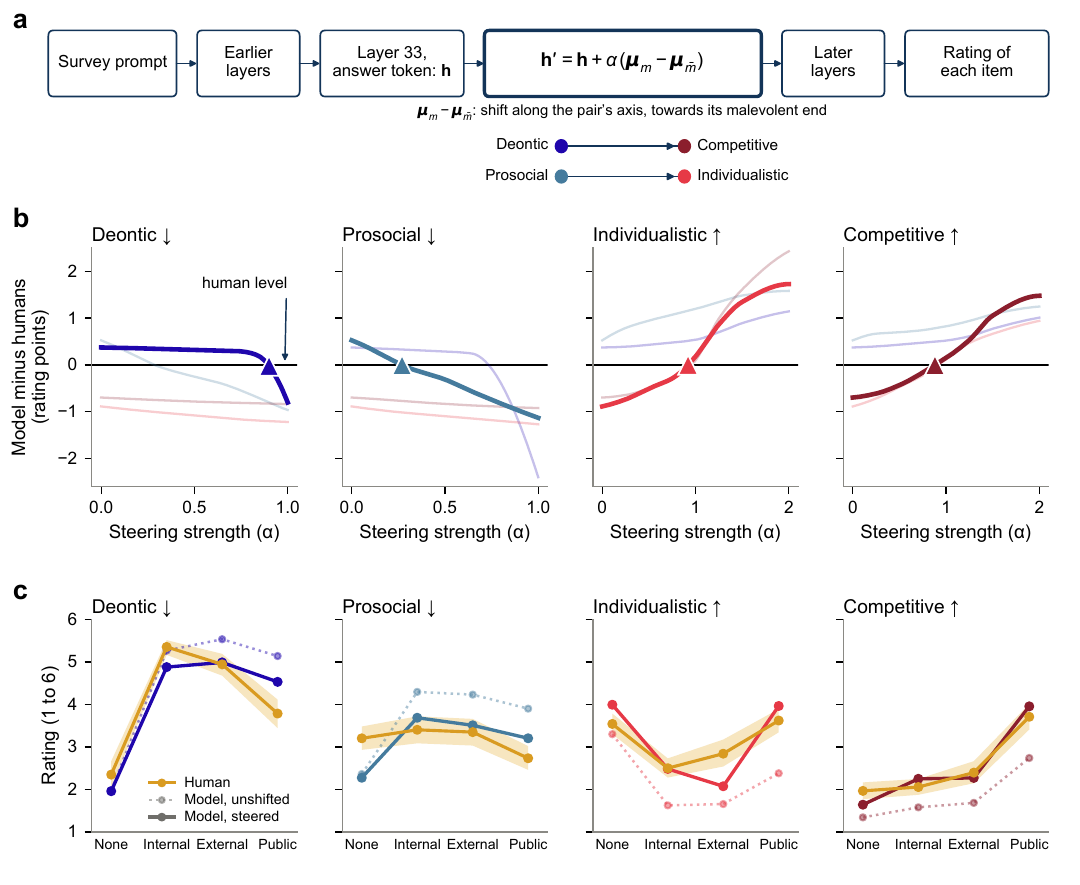}
\caption{\textbf{Moving each motive along its own axis.} Gemma~4 12B instruction-tuned, layer 33, all 12 items, 50 simulated respondents per condition in each survey version; each version is compared with its own human sample (Brotzeller et al.\ Studies~2 and~3) and the two are averaged. (a) At the answer token, $\alpha$ times the difference between the mean activation of the corpus texts stating motive $m$, $\boldsymbol{\mu}_m$, and that of the texts stating its opposite, $\boldsymbol{\mu}_{\bar m}$, is added to the residual stream $\mathbf{h}$; $\alpha = 1$ moves the activation by the whole difference. The opposites are the pairs of the social value orientation framework; a malevolent motive is pushed towards itself ($\uparrow$), a benevolent motive towards its opposite ($\downarrow$). (b) For each steered motive, the model's rating minus the human rating, averaged over the four conditions, for all four motives against $\alpha$: the steered motive in full colour, the others faint (deontic indigo, prosocial blue, individualistic red, competitive dark red). Curves interpolate the measured strengths (Methods); the triangle marks the measured strength at which the steered motive is closest to the human level (deontic 0.90, prosocial 0.27, individualistic 0.92, competitive 0.88). (c) Rating of each motive by condition for humans (gold line and 95\% band), the unshifted model (dotted) and the model steered along that motive's axis at the strength of the triangle (solid); model intervals over simulated respondents are narrower than the markers. No setting degraded the output (digit mass 1.00).}
\end{figure}

\section{Discussion}\label{sec:discussion}

Aligned language models attribute more benevolent and less malevolent motive
to a whistleblower than humans do, and the two halves of that divergence
have different origins. On the malevolent side the model reports what it
represents. In a preregistered confirmatory run on new simulated respondents,
the internal and the expressed reading of the individualistic motive did not
differ, and neither did those of the competitive motive. Their difference
was smaller than the shift that a single sentence giving the whistleblower a
grudge produced in both readings. On the benevolent side the ratings run ahead of
the representation. The expressed deontic and prosocial readings exceeded the
internal ones by 0.5 to 0.7 of the anchor spacing in the two registered
conditions and by 0.2 to 0.6 for public disclosure, and the deontic rating of the original scenario
exceeded the rating the model gave to texts in which duty is stated outright,
whereas the representation kept the two apart. The missing malice is missing
from the probed representation; the surplus virtue is added at the report.

The two halves also respond differently when the representation is moved: shifts
along either malevolent axis and along the prosocial axis bring the ratings smoothly to
the human level without degrading the output, whereas the deontic surplus yields only
past a threshold, at a point where the prosocial rating is already far below the human
one. Because each shift also moves the other motives, no single shift reproduces the
human profile. Where the deontic surplus is set, whether at the readout, in the policy
learned at post-training or elsewhere, the present measurements do not say.

The pattern is consistent with a model that is charitable by default. Humans
attribute some self-interest and some intent to harm to a whistleblower, more
when he goes to the press, and the models reproduce that ordering at a lower
level. In free text the difference is larger: humans named a competitive motive in
one answer in ten, the model in at most one in a hundred, and the model produced
the attribution readily once the scenario supplied a reason for it. The
model can make the attribution; what the original scenario lacks, for the model, is a reason to form it. On
the benevolent side the models go beyond charity: they report more duty and
more concern for others than they represent, and more than humans report.
For anyone who lets a language model weigh another person's conduct, and for
researchers who use models as simulated participants, the practical
consequence is that the two biases need different remedies. A reporting bias
can in principle be addressed at the output, by elicitation, decoding or
calibration against human data. A weak internal signal cannot be addressed
in this way, and a remedy that made the model say more about malice without changing what it
represents would produce the pattern we found on the benevolent side.

Four features of the measurement bear on how far these conclusions reach.
First, the probe reads a linear direction defined through a corpus of stated
motives. The admission checks, the explicit versions of the scenario and the
cue sentences establish that it is not keyed to a word or a frame and that it
reads stated and implied motive; they do not establish that it captures everything the model
represents about motive. An internal reading near zero is therefore the absence of
a linear signal on this direction, not the absence of all relevant information.
Second, the scale on which the two readings are compared depends on where its
zero is placed. The registered pairing places it at the motive's opposite in
the social value orientation framework. The first preregistration paired the
motives differently, and a third scheme places the zero at the other three
motives together. The benevolent result is the same under all three, and the
competitive difference stays small under all three, $-0.08$ to $-0.20$ in the
two registered conditions (Extended Data Fig.~3b). Third, the corpus was written by the authors
with language-model assistance, every event appearing in all four
motives on an identical background. It was screened by five language-model raters and
inspected internally for clarity and consistency (Methods, ``Probe
corpus'').
Fourth, the mechanism is read in one open model of 12 billion parameters.
The behavioural profile it explains appears in four frontier models of
another family and in the five models of our earlier work (Fig.~1c), but this method cannot
establish whether the mechanism is the same in a model whose weights cannot be read.

The clinical literature describes the over-attribution of intent to harm,
in aggressive children and in paranoia \citep{CrickDodge1994, Combs2007},
and the measurement framework used here applies equally to under-attribution
and over-attribution. Language models pass
many explicit theory-of-mind tests \citep{Kosinski2024, Strachan2024}, but
success on such tests does not guarantee robust human-like mental-state
inference \citep{Shapira2024}. The present result shows that a model can
match human verdicts while differing from them in a spontaneous attribution
the tests do not ask about, and in two different ways at once. A simulated
respondent that does not infer malice where humans do, and reports virtue
beyond what it infers, is not an unbiased observer. It has a
different attributional prior, one that, like the assumption that people
choose more rationally than they do \citep{liu2024large}, idealises the people
it is asked to simulate. The same tools that locate the two halves of the
prior in a model could be used to ask where each is acquired.

\section{Methods}\label{sec:methods}

\subsection{Models}
We used the instruction-tuned release of Gemma~4 12B \citep{GemmaTeam2026}
(\texttt{google/\allowbreak{}gemma-\allowbreak{}4-\allowbreak{}12B-\allowbreak{}it}; 48 decoder layers), run in bfloat16 with
Hugging Face \texttt{transformers} \citep{Wolf2020} on a single RTX~5090 (32~GB), with no
sampling and no generated text; its thinking mode was not used. The whole
prompt was passed through the model's chat template as a single user message.
The matched pre-trained release (\texttt{google/\allowbreak{}gemma-\allowbreak{}4-\allowbreak{}12B}) was run in
parallel on the identical text with no template, in the pilot run and again
in the confirmatory run. Its rating scale was compressed
(corpus anchors 0.45 to 0.50 scale points apart against 2.32 to 2.71 in the
confirmatory run), so the instruction-tuned minus pre-trained contrast
registered in the first study (H2) was uninformative. Its internal readings of the competitive motive were no higher
than those of the instruction-tuned weights ($-0.40$ to $-0.31$ in the two
layers admitted for it in the confirmatory run), which agrees with the
conclusion reported here. All other analyses used the instruction-tuned
weights.

\subsection{Scenario, conditions and respondents}
The scenario is the whistleblowing vignette of Brotzeller et al.\
\citep{Brotzeller2025}, Study~2, as used in our earlier work \citep{WuDreher2026}.
The respondent is told to imagine being a physician at a clinic where a
colleague, Dr Schmitt, bills the insurer for services not provided, and that
another colleague, Dr Bauer, learns of this and either passes the information to
no one (no disclosure), to the clinic's compliance office (internal), to the
authorities (external) or to a national newspaper (public). Each simulated
respondent was assigned a persona (age, gender and student status, sampled
with the same distributions as the human sample of Brotzeller et al.\
\citep{Brotzeller2025}), and the genders of the two
physicians were sampled per respondent. No new human-participant data were
collected for this study. The human
ratings quoted in the Results come from the two human samples of our earlier
work \citep{WuDreher2026}, computed from the published de-identified data of
Brotzeller et al.\ \citep{Brotzeller2025} (``Motive profile'' below), and
the open answers come from Study~1 of the same authors (``Free-text
completions''). Personas were drawn from a fixed seed per run, so that within a
run every model and both weights rate the same 200 respondents per condition;
the pilot, second-hand and Claude runs share one set of personas, and the
confirmatory run drew a new set.

\subsection{Items and versions of the scenario}
The four motives were measured with the 12 items of the original
instrument, three per motive, each rated on a six-point scale from 1
(strongly disagree) to 6 (strongly agree): deontic (``Dr Bauer wanted to
follow a moral obligation'', ``\ldots act in accordance with moral values'',
``\ldots do the right thing''), prosocial (``\ldots help others'', ``\ldots
support others'', ``\ldots assist others''), individualistic (``\ldots benefit
themselves'', ``\ldots pursue their own interests'', ``\ldots gain an advantage
for themselves'') and competitive (``\ldots harm others'', ``\ldots humiliate
others'', ``\ldots make others look bad''). We use the category names of
Brotzeller et al.\ \citep{Brotzeller2025} throughout. Each respondent rated
each item in a separate prompt. Three versions of every scenario were run for
every motive: the original; an explicit version with a sentence stating the
motive appended (``Dr Bauer did this because he believed it was his
professional duty'', ``\ldots because he wanted to protect the patients'',
``\ldots because he hoped it would advance his own career'', ``\ldots because
he wanted to harm Dr Schmitt'', pronouns matched to the sampled gender); and a
neutral version with ``Dr Bauer did this on a Tuesday afternoon''. For the
competitive motive a negated version (``Dr Bauer did this although he did not
want to harm Dr Schmitt'') was run as well. The original version provides
the measurements, the neutral and negated versions the validity
checks, and the explicit versions the manipulation checks. Per weight this
gives $200 \times 4 \times 12 \times 3 + 200 \times 4 \times 3 = 31{,}200$
scenario prompts. The first study used the three competitive
items with an explicitly competitive and an explicitly protective version, as described in its
published analysis; the runs reported there are
released with the results.

\subsection{Motive profile}
For the comparison with humans (Fig.~1 and Extended Data Fig.~1) the
expressed ratings of the 12 items in the original version of the
confirmatory run give the first-hand profile ($200 \times 4 \times 12 =
9{,}600$ prompts). The second-hand profile comes from a run of the pilot
personas on the second-hand version of the scenario (Brotzeller et al.\
Study~3), in which the events are introduced as something the respondent
``happened to hear about'' that ``happened on your team some time ago'' and
the decision as ``You also learned that one of your colleagues, Dr Bauer,
learned about it and subsequently \ldots''. The wording is that of our
earlier work \citep{WuDreher2026}. No rating fell below the digit-mass
threshold in either run. A respondent's score for a motive is the mean
expected rating over its three items. The human values are condition means of
the per-participant motive scores in the published data of Study~2 and
Study~3 \citep{Brotzeller2025}, after the exclusions of our earlier work
(participants were retained only if they consented to the use of the data and
answered the attention check and both comprehension questions correctly), leaving 125 participants (30, 33, 30 and 32
per condition) and 742 participants (179, 189, 190 and 184). Intervals are
95\% bootstrap intervals over participants. Fig.~1 averages the two
perspectives with equal weight, and the interval of the human average is the
average of the two intervals.

\subsection{Frontier models}
Claude models (Fig.~1b). Claude Haiku~5.5 (\texttt{claude-\allowbreak{}haiku-\allowbreak{}5-\allowbreak{}5}), Sonnet~5.5
(\texttt{claude-\allowbreak{}sonnet-\allowbreak{}5-\allowbreak{}5}), Opus~5.5 (\texttt{claude-\allowbreak{}opus-\allowbreak{}5-\allowbreak{}5}) and Fable~5.1
(\texttt{claude-\allowbreak{}fable-\allowbreak{}5-\allowbreak{}1}) \citep{Anthropic2026cards, Anthropic2026sonnet, Anthropic2026fable} were run through the Anthropic Messages API on 20
and 21 October 2026. Each request contained one survey prompt of the Gemma
runs (Supplementary Methods) as a single user message, without a system
message, for the 200 pilot personas per condition, 12 items and two
versions of the scenario (9,600 requests per model and version). Requests were sent at temperature 1
with one sampled answer, and the rating was read as the digit in the reply. Thinking was
disabled for Haiku and Sonnet, Fable answered without thinking (maximum 5
output tokens for these three), and Opus, for which thinking cannot be
disabled, used adaptive thinking at effort low (maximum 1,024 output tokens).
A reply without a digit from 1 to 6 counted as missing (one reply each in two
runs). A respondent's score for a motive is the mean over its three items.

Models of the earlier study (Fig.~1c). The ratings of GPT-4o
(\texttt{gpt-\allowbreak{}4o-\allowbreak{}2024-\allowbreak{}08-\allowbreak{}06}, OpenAI API), gpt-oss-120b (Cerebras API), Gemini~2.5
Pro (\texttt{gemini-\allowbreak{}2.5-\allowbreak{}pro}, Gemini API), DeepSeek~V3 (\texttt{deepseek-\allowbreak{}chat},
DeepSeek API) and Llama~3.1 8B Instruct
(\texttt{meta-\allowbreak{}llama/\allowbreak{}Llama-\allowbreak{}3.1-\allowbreak{}8B-\allowbreak{}Instruct}, Hugging Face Inference API) were
collected from 21 to 24 June 2026 and reported in our earlier study
\citep{WuDreher2026}, and are
re-plotted here from the data of that study. Each simulated respondent
received one request, consisting of a system message with the persona (age, gender and
student status sampled from the demographics of the corresponding human
sample) and a user message with the scenario, the 12 motive items
and 16 character items in random order and a JSON answer format. Sampling was
at temperature 0.8, with 200 respondents per condition and version of the scenario.
Scenario, items and scale are those of Brotzeller et al. Prompts, run dates, maximum output length and
reasoning setting of every model are listed in the Supplementary Methods and
Supplementary Table~7.

\subsection{Prompt and readout}
Every prompt consisted of the persona paragraph, the scenario, a reminder
sentence naming the person the statement is about, the six-point scale, the
instruction ``Answer with a single number from 1 to 6 and nothing else'', the
statement, and the string ``Answer:'' (the complete prompts are given verbatim in the Supplementary Methods). For pre-trained weights the single-number
instruction was omitted, four example ratings from an unrelated survey (a true
statement rated 6, an absurd one rated 1, an everyday opinion rated 4, and a
statement attributing an annoying motive to a neighbour rated 2) were
inserted after the persona paragraph, the prompt ended with ``Rating (1-6):'',
and a space token was appended, because digits are tokenised without a leading
space and a pre-trained model emits the space before the digit.
The prompt style of each weight was fixed on the basis of a readout check run before any scenario
analysis (``Readout check'' below). Let $p(k \mid x)$ be the model's next-token probability at
the answer position of prompt $x$ for the digit $k \in \{1, \dots, 6\}$, summed
over all vocabulary tokens that decode to that digit. The digit mass $M(x)$, the
renormalised digit distribution $\tilde p_k(x)$ and the expressed rating $R(x)$
are
\begin{equation}
M(x) = \sum_{k=1}^{6} p(k \mid x), \qquad
\tilde p_k(x) = \frac{p(k \mid x)}{M(x)}, \qquad
R(x) = \sum_{k=1}^{6} k \, \tilde p_k(x),
\label{eq:rating}
\end{equation}
so that $R(x)$ is the probability-weighted mean rating on the 1 to 6 scale and
$M(x)$ measures how much of the model's prediction is a rating at all. Any
rating with $M(x) < 0.20$ was excluded from all averages, as preregistered. In the 200-respondent runs no rating of the original scenario
was excluded in any model (the explicit prosocial version with external disclosure produced no digit; see ``Positive controls''). In the first study, of the 1,260 corpus ratings per run, 23
(Gemma~4 instruction-tuned), 40 (Gemma~4 pre-trained), 28 and 37 (the same
weights under the second wording) were excluded.

\subsection{Probe corpus}
The probe corpus consists of 70 everyday events, each written as a short text
with a shared background (stating every fact the four motive sentences rely on
and nothing else), an action, and one of four explicitly stated motives:
deontic, prosocial, individualistic or competitive (280 texts; mean 59 words). The
events are everyday situations in sport, school, family, work, neighbourhood,
friendship and online settings, chosen to be unlike the medical billing fraud of
the scenario. Fifty are reports of some kind (telling a coach, a landlord, a
manager, a moderator) and 20 are not reports (helping, lending, declining,
correcting). Because every event appears with all four motives, the action is
uninformative about the motive. Before use,
the 280 texts were validated by five language-model raters (GPT-5.6 at
four reasoning settings and DeepSeek V4.1 Flash), which received the texts with
the labels removed and the order shuffled and rated each text on the four
motive dimensions (1 to 6). The written motive received mean ratings of 5.4 to
6.0 and the other three dimensions 1.1 to 1.4 on average; competitive ratings of
the individualistic texts, the closest confusion, were 1.1 to 2.3. The
top-rated dimension matched the written motive for 99.6 to 100\% of the texts
in four raters and 86\% in the fifth, which left 16 texts unrated; competitive
ratings correlated 0.92 to 0.99 between raters. Two texts flagged by this
check were reworded before the corpus was used. For every
competitive text two further texts were generated: a negated version in which the
competitive motive sentence is denied (``She did not want X''; labelled not
competitive) and a contrastive version in which a benign motive is denied and the
competitive one asserted (``She did not do it out of concern for anyone. She wanted
X''; labelled competitive), giving 420 texts. Each corpus text was placed in the
same prompt frame as the scenario, with the actor's name in the reminder sentence and
in the statement, and rated on the three items of each motive, giving 1,260 corpus
prompts per motive and weight.

Human validation. Three volunteers not involved in the study also judged
the motive of each text, in a five-way forced choice (the four motives or
``cannot tell''), with the instruction to judge from the sentence stating
why the person acted. Two raters judged 210 texts each and one 140, so
that every text was judged by exactly two raters (560 judgements). Agreement with the corpus
label was 98\% (Cohen's $\kappa = 0.97$ \citep{Cohen1960}); both raters agreed with the label
for 96\% of the texts and at least one did for every text. The corpus
was not altered after this check, which was completed after registration; the
judgements are released per rater and per text with the data.

\subsection{Probe training and layer selection}
At the answer position of every prompt, the residual-stream activation
after every third decoder layer from layer 6 to the last multiple of three was stored. We
write $\mathbf{h}_\ell(x) \in \mathbb{R}^{d}$ for the output of decoder block
$\ell$ at the final token of prompt $x$, the position at which the rating digit
is predicted, where $d$ is the model's hidden size. For each motive $m$ and
layer a logistic regression ($L_2$ penalty, $C = 0.5$; classes weighted; scikit-learn \citep{Pedregosa2011}) was
trained to separate the corpus texts stating $m$ (for the competitive motive,
the competitive and the contrastive texts) from all other corpus texts (the
texts stating the three other motives and, for the competitive probe, the
negated texts), with five-fold cross-validation grouped by event so that every
event is unseen when it is scored. The probe acts on standardised activations,
and its decision value for a prompt is
\begin{equation}
s_\ell(x) = \mathbf{w}_\ell^{\top} \mathbf{z}_\ell(x) + b_\ell, \qquad
z_{\ell,i}(x) = \frac{h_{\ell,i}(x) - \bar h_{\ell,i}}{\sigma_{\ell,i}},
\label{eq:probe}
\end{equation}
where $\bar h_{\ell,i}$ and $\sigma_{\ell,i}$ are the mean and standard deviation of
dimension $i$ over the training activations, and $\mathbf{w}_\ell$ and $b_\ell$
are the fitted weights and intercept; $s_\ell(x) > 0$ assigns the prompt to
motive $m$. Probe accuracy is the mean held-out area under the receiver operating characteristic curve (ROC AUC) across folds; for the
competitive probe the held-out probability assigned to the negated texts is
reported as a lexical check. The probe used on the scenario was refit on all
texts.

A layer was admitted to the analysis of motive $m$ if
\begin{equation}
\begin{gathered}
\mathrm{AUC}_\ell \ge 0.8, \qquad
\max_{c} \bigl| \bar I^{\,\ell}_{c,\mathrm{neutral}} - \bar I^{\,\ell}_{c,\mathrm{original}} \bigr| < 0.2, \\
\max_{c} \bar I^{\,\ell}_{c,\mathrm{negated}} < 0.5 \quad \text{(competitive probe only)},
\end{gathered}
\label{eq:admit}
\end{equation}
where $\mathrm{AUC}_\ell$ is the held-out probe accuracy and
$\bar I^{\,\ell}_{c,v}$ is the mean internal index (equation~\eqref{eq:int})
at layer $\ell$ of version $v$ of the scenario in disclosure condition $c$,
averaged over respondents and items. The conditions require, in turn, that
the probe separates the motive's texts from the others on unseen events, that
an irrelevant sentence does not move the reading, and, for the competitive
probe, that a sentence denying the motive is not read as the motive. The
analysis window of a weight for a motive is the set of its own admitted
layers: in the instruction-tuned weights of Gemma~4, layers 15, 18 and 27 to
45 for the deontic motive, 33 to 45 for the prosocial motive and 27 to 45 for
the individualistic and the competitive motive; in the pre-trained weights,
layers 18 and 21 for the deontic and 27 and 45 for the competitive motive,
with no admitted layer for the other two. The instruction-tuned minus
pre-trained contrast (H2) is computed in the layers
admitted in both weights. The windows were fixed in the pilot run and
deposited with the frozen code at registration; every layer passed the checks
again in the confirmatory data. The per-layer values of every motive, and those of the first
study, are released with the results.

\subsection{Index and difference}
For each layer the decision function of motive $m$'s probe (equation~\eqref{eq:probe}) was applied to every corpus and scenario prompt. Let $o(m)$ be the paired
opposite of $m$ in the social value orientation framework, competitive for
deontic, individualistic for prosocial and vice versa, and let
$\mathcal{C}_m$ and $\mathcal{C}_{o(m)}$ be the corpus prompts built from the
70 texts stating $m$ and the 70 stating $o(m)$ (three items each, 210 prompts
per set; the negated and contrastive texts are not used as anchors). The
internal anchors at layer $\ell$ are the means of the out-of-fold decision
values $\tilde s^{m}_\ell$ from the grouped cross-validation,
\begin{equation}
\mu^{\ell}_{m,0} = \frac{1}{|\mathcal{C}_{o(m)}|} \sum_{x \in \mathcal{C}_{o(m)}} \tilde s^{m}_\ell(x), \qquad
\mu^{\ell}_{m,1} = \frac{1}{|\mathcal{C}_{m}|} \sum_{x \in \mathcal{C}_{m}} \tilde s^{m}_\ell(x),
\label{eq:anchors}
\end{equation}
where $\tilde s^{m}_\ell(x)$ is the decision value of the fold model for which
the event of $x$ was held out. A scenario prompt for an item of motive $m$,
out of sample for every fold, is scored by the probe refit on all 420 texts,
and its internal index at layer $\ell$ is
\begin{equation}
I^{\ell}_{\mathrm{int}}(x) = \frac{s^{m}_\ell(x) - \mu^{\ell}_{m,0}}{\mu^{\ell}_{m,1} - \mu^{\ell}_{m,0}}.
\label{eq:int}
\end{equation}
The expressed index applies the same transformation to the rating of
equation~\eqref{eq:rating}, with anchors taken from the ratings of the same
corpus prompts on the items of motive $m$,
\begin{equation}
I_{\mathrm{exp}}(x) = \frac{R(x) - \bar R_{m,0}}{\bar R_{m,1} - \bar R_{m,0}}, \qquad
\bar R_{m,0} = \frac{1}{|\mathcal{C}'_{o(m)}|} \sum_{x \in \mathcal{C}'_{o(m)}} R(x), \quad
\bar R_{m,1} = \frac{1}{|\mathcal{C}'_{m}|} \sum_{x \in \mathcal{C}'_{m}} R(x),
\label{eq:exp}
\end{equation}
where the primed sets contain the prompts with $M(x) \ge 0.20$. On both
indices the texts stating the opposite motive sit at 0 and the texts stating
the motive at 1. The denominator $\bar R_{m,1} - \bar R_{m,0}$ is the anchor
spacing of the rating scale; an expressed scale is used only if it is at
least 0.3 scale points (as preregistered). The first preregistration paired
the motives differently (competitive against prosocial, deontic against
individualistic); a third scheme places the zero at the texts of the other
three motives. All three are linear rescalings of the same readings; the
registered scheme is used in the Results and the other two are reported in
Extended Data Fig.~3b.

For respondent $r$, condition $c$ and motive $m$, let $J_{rcm}$ be the items
of $m$ whose rating has $M \ge 0.20$ (all three in every run reported) and
$W_m$ the analysis window of the motive. The internal and expressed readings
of the respondent are averaged over items and, for the internal reading, over
the window, and the quantity tested is their difference, expressed minus
internal:
\begin{equation}
\begin{aligned}
I_{\mathrm{int}}(r,c,m) &= \frac{1}{|J_{rcm}|} \sum_{j \in J_{rcm}} \frac{1}{|W_m|} \sum_{\ell \in W_m} I^{\ell}_{\mathrm{int}}(x_{rcj}), \\
I_{\mathrm{exp}}(r,c,m) &= \frac{1}{|J_{rcm}|} \sum_{j \in J_{rcm}} I_{\mathrm{exp}}(x_{rcj}), \\
D(r,c,m) &= I_{\mathrm{exp}}(r,c,m) - I_{\mathrm{int}}(r,c,m).
\end{aligned}
\label{eq:gap}
\end{equation}
The condition values reported are the means over the $N = 200$ respondents.
A positive $\bar D$ means that the model reports more of the motive than it
represents, a negative $\bar D$ that it represents more than it reports (the
direction suppression predicts; the gap $G = -D$ of the first
preregistration), and zero that the two readings agree. The registered
decisions are taken in the internal and external disclosure
conditions: for the deontic and prosocial motive, intervals above zero in
both support H1b and intervals including zero in both support H1a; for the
individualistic and competitive motive, intervals below zero in both support
H2b and intervals including zero in both support H2a; a disagreement between
the two conditions supports neither account. The instruction-tuned minus
pre-trained contrast (H2) is reported for the
competitive motive under ``Models'' and is not used for the
decisions above.

Two sources of uncertainty are reported, both as percentile bootstrap
intervals \citep{EfronTibshirani1993}. Respondent-level
intervals (4,000 resamples over respondents) are narrow ($\pm 0.01$ to $\pm 0.02$)
because the readout is deterministic given the prompt and personas contribute
little variance (the 200-respondent values reproduced a 20-respondent pilot
to within 0.03). They are given in the released data. The intervals in the
text and figures are event-level, because the probe and all four anchors are
estimated from 70 events. In each of $B$ resamples $b$, the 70 corpus
events were drawn with replacement; the probe of equation~\eqref{eq:probe} was
refit at every window layer on the resampled corpus, the anchors of
equations~\eqref{eq:anchors} and~\eqref{eq:exp} were recomputed from it, and
$\bar I^{(b)}_{\mathrm{int},c}$, $\bar I^{(b)}_{\mathrm{exp},c}$ and
$\bar D^{(b)}_c$ were recomputed from equations~\eqref{eq:int}
to~\eqref{eq:gap} for the unchanged scenario prompts. The interval is the
2.5th to 97.5th percentile of the $B$ values. The same event resamples were
drawn for the instruction-tuned and the pre-trained weights, so the instruction-tuned minus pre-trained contrast (H2) is taken from the paired
resamples. For the confirmatory run $B = 10{,}000$; for the bands of Fig.~2c, $B = 200$. Under the registered rule, a difference is called
positive or negative only when its event-level interval excludes zero. The
two-sided bootstrap $p$ value, $p = 2\min\{(1 + \#\{b: \bar D^{(b)}_c \le 0\})/(B + 1),\ (1 + \#\{b: \bar D^{(b)}_c \ge 0\})/(B + 1)\}$
\citep{PhipsonSmyth2010}, Bonferroni-corrected over the 16 motive by condition cells, is also reported (Extended Data Table~2).

\subsection{Readout check}
Before any scenario analysis each weight was run on a diagnostic set: a true
statement (``Dr Bauer is a physician'') and an absurd one (``Dr Bauer is a
fish''), the original scenario, and its explicitly competitive and explicitly protective
control versions, under each prompt style (plain, single-number instruction,
four-example), for one fixed persona in the public-disclosure condition. The
style with the widest separation between the two control versions and
adequate digit mass was adopted (single-number instruction for all
instruction-tuned weights, four-example with a trailing space for all
pre-trained weights). The values reported here come from a repetition of the check in the exact prompt frame of the analysis runs; the original check, run with a longer reminder sentence, is in the released data and leads to the same choices. In the adopted styles the true
and the absurd statement were separated by 3.90 scale points in Gemma~4 12B
instruction-tuned (4.90 against 1.00) and 3.18 in its pre-trained weights (5.48
against 2.30); the explicitly competitive and explicitly protective versions were
separated by 1.05 scale points in the instruction-tuned weights (2.06 against
1.01) and 1.03 in the pre-trained weights (2.84 against 1.81), with digit mass
of 1.00 in the instruction-tuned and 0.73 to 0.91 in the pre-trained weights.
An expressed scale was used only if the two corpus
anchors obtained in the full run were at least 0.3 scale points apart, a
criterion fixed after this check and before any gap was examined (the raw ratings and anchors of the first study are released with the results).

\subsection{Intervention on the representation}
For Gemma~4 12B the residual stream was modified at the answer token of the
original scenario prompts (20 respondents per condition, the three competitive
items plus one individualistic and one prosocial item) at one decoder layer
$\ell$, and the rating was read as above. The intervention adds a fixed vector to the
output of block $\ell$ at the final token only, leaving every other position
unchanged:
\begin{equation}
\mathbf{h}'_\ell(x) = \mathbf{h}_\ell(x) + \alpha \, \delta \, \mathbf{u},
\label{eq:steer}
\end{equation}
where $\mathbf{u}$ is a unit direction in raw activation units, $\delta$ a scale
in the same units and $\alpha$ the size of the shift. All directions were
computed from the stored corpus activations at that layer. With
$\boldsymbol{\mu}_m$ the mean activation $\mathbf{h}_\ell$ over the corpus
prompts with stated motive $m$ (three items each), the main direction is the
competitive-minus-prosocial difference of corpus means,
\begin{equation}
\mathbf{u}_H = \frac{\boldsymbol{\mu}_H - \boldsymbol{\mu}_P}{\lVert \boldsymbol{\mu}_H - \boldsymbol{\mu}_P \rVert}, \qquad
\delta_H = \lVert \boldsymbol{\mu}_H - \boldsymbol{\mu}_P \rVert,
\label{eq:dir}
\end{equation}
so that $\alpha = 1$ adds exactly the difference between the competitive and the
prosocial corpus means ($\delta_H = 20.6$ at layer 30). Two further directions
were used: the probe weight vector in raw units,
$\mathbf{u}_w \propto \mathbf{w}_\ell \oslash \boldsymbol{\sigma}_\ell$
(equation~\eqref{eq:probe}), and, as a control, the deontic-minus-individualistic
difference of corpus means,
$\mathbf{u}_C \propto \boldsymbol{\mu}_D - \boldsymbol{\mu}_S$, used both as is
and after removing its component along $\mathbf{u}_H$,
$\mathbf{u}_{C\perp} \propto \mathbf{u}_C - (\mathbf{u}_C^{\top}\mathbf{u}_H)\,\mathbf{u}_H$
(each normalised to unit length). For a direction $\mathbf{u}$ with its own
motive contrast $(m_+, m_-)$, the natural scale is the distance between the two
anchor means projected onto it, $\delta = \mathbf{u}^{\top}(\boldsymbol{\mu}_{m_+} - \boldsymbol{\mu}_{m_-})$,
which equals $\delta_H$ for $\mathbf{u}_H$. The shifted ratings were converted
to the expressed index with the unshifted corpus anchors of
equation~\eqref{eq:exp}. The shifts were $\alpha \in \{-1, -0.5, 0.5, 1\}$, plus
$\pm 2$ at layer 30 for $\mathbf{u}_H$ and up to $\pm 20$ for
$\mathbf{u}_w$. The probe direction proved unusable as an intervention. It concentrates on
low-variance dimensions, and the two corpus anchors are only 0.07 units apart
along it; shifts of up to five times that
distance moved the rating by at most 0.07 of the anchor spacing, and shifts
of 10 to 20 times it destroyed the model's ability to answer with a
digit (digit mass 0 at $+10$, $+20$ and $-20$). Along the
competitive-minus-prosocial direction, $\alpha = \pm 2$ lost the item
specificity seen at $|\alpha| \leq 1$ (at $-2$ the individualistic and
prosocial items fell to 1.34 and 1.26; at $+2$ the individualistic item rose to
2.60), so the item-specific range of the intervention is $|\alpha| \leq 1$.
The two control directions were run twice: once
with their own scale, $\delta = \mathbf{u}^{\top}(\boldsymbol{\mu}_D - \boldsymbol{\mu}_S)$
(4.4 units for $\mathbf{u}_{C\perp}$, a fifth of $\delta_H$), and once with
$\delta = \delta_H$, so that for every $\alpha$ the control shift has the same
norm as the competitive shift; the equal-norm runs are the ones reported (Extended Data Fig.~5). Five
random unit directions, drawn from an isotropic Gaussian and normalised, were
run with $\delta = \delta_H$ as a further control. A setting in which more than 20\%
of ratings had digit mass below 0.2 was treated as degraded and not
interpreted. Layers 24, 27, 30, 33, 36 and 39 were run for the
competitive-minus-prosocial direction; layer 30 was used for the controls and
for the pre-trained weights. The mean digit mass over the five random directions was 0.03 to 0.05 at $|\alpha| = 1$. The full trajectories, which Extended Data Fig.~5 shows, are released with the results.

Steering each motive. For Fig.~5 the intervention $\mathbf{h}' = \mathbf{h} +
\alpha(\boldsymbol{\mu}_m - \boldsymbol{\mu}_{\bar m})$ was applied at layer 33 of the instruction-tuned weights,
the first layer in the analysis window of all four motives, and at the answer token
only. For motive $m$, $\boldsymbol{\mu}_m$ is the mean activation of the corpus texts
stating $m$ and $\boldsymbol{\mu}_{\bar m}$ that of the texts stating its opposite in
the social value orientation pairing (deontic against competitive, prosocial against
individualistic), both computed on the corpus prompts presented with $m$'s three survey
items; the length of the difference is 44.7, 44.9, 34.6 and 40.4 raw units for the
competitive, individualistic, deontic and prosocial motive. These directions are
differences of class means, the axis of the
motive index, not the probes' weight vectors. Malevolent motives were shifted towards
the motive on the grid $\alpha \in \{0, 0.25, 0.5, 0.75, 1, 1.25, 1.5, 2\}$ and benevolent motives
away from it on $\alpha \in \{0, 0.05, 0.1, 0.15, 0.2, 0.3, 0.5, 1\}$. The grid was then
refined around the point at which the steered motive's bias relative to humans, averaged over the four conditions
and the two versions, changed sign: at 0.88 and 0.91 for the competitive motive, 0.27 and 0.64 for the prosocial motive,
0.88 and 0.91 to 1.00 in steps of 0.01 for the individualistic motive, and 0.70 to 0.95 in steps of 0.05 and then 0.86 to 0.89
in steps of 0.01 for the deontic motive. The strength reported is the measured point at
which that bias is closest to zero. The curves of Fig.~5b are shape-preserving cubic
interpolations through the measured strengths. Each setting rated all 12 items for 50 simulated
respondents per condition in both survey versions, each version compared with
its own human sample (first-hand, Study~2; second-hand, Study~3). The bias is the model
mean minus the human mean. A setting in which more than 20\% of ratings had digit
mass below 0.2 would have been treated as degraded; none was (digit mass 1.00
throughout). A repetition at the selected strengths with new simulated respondents
reproduced every motive by condition mean, averaged over the two versions, to within 0.04.

\subsection{Positive controls}
Three analyses test whether the method detects motives that are present.
(1) Explicit versions. Each motive's explicit version of the scenario
(``Items and versions of the scenario'') is read by the motive's probe and rated on its
items; the probe must read it at 0.5 or above. In the confirmatory run the
window means were 0.78 to 0.92 (deontic), 0.57 to
0.68 (individualistic) and 0.90 to 1.01 (competitive) over the four conditions, and 0.67 and 0.78 for the prosocial
motive with internal and public disclosure; the explicit prosocial version with no disclosure, a condition in which no
hypothesis is decided, was read at 0.23. The explicit
ratings were at or near the level of the corpus texts stating the motive
(expressed index 0.87 to 1.44) except for the competitive motive (0.52 to
1.02). The explicit
prosocial version with external disclosure produced no digit (digit mass 0
for all 600 prompts, in the pilot and the confirmatory run alike) and is
excluded from this check. (2) Cue sentences (Fig.~3). One sentence was
inserted before the ending of the scenario, with pronouns matched to the
sampled genders (the four sentences are shown in Fig.~3a). Each cue version was read on all 12 items with the
respondents of the pilot run, using the probes, anchors and layer windows of the
first preregistration's pairing (zero at the deontic texts for the individualistic
motive and conversely, and at the prosocial texts for the competitive motive and
conversely; windows 27 to 45 for the deontic, individualistic and competitive
probes and 24 and 33 to 45 for the prosocial probe), under which the cue runs were
made before the second registration. The change
from the original scenario in the internal and the expressed reading is
reported as the mean over the internal, external and public disclosure
conditions with 95\% event-bootstrap intervals. Human participants never saw these
sentences; they are controls on the measurement, not conditions of the
comparison with humans. The cue sentences were written to imply one motive each; their effects
on all four motives are reported (Fig.~3b). (3) Implied competitive motive, from the first study. The 70
corpus events were rewritten as background and action only and read through the
competitive probe; the 17 backgrounds that give the
actor a reason for resentment read 0.19 $[0.11, 0.28]$ above the others
internally and 0.22 $[0.11, 0.34]$ in the rating.

\subsection{Free-text completions}
Study~1 of Brotzeller et al.\ \citep{Brotzeller2025} asked 201 participants,
in German, why an insurance-company employee, Stephan or Stephanie, who
learned that colleagues had given customers misleading advice, kept quiet or
reported it internally, to the financial supervisory authority or to a
newspaper. The open answers are available under a scientific use licence
(PsychArchives, doi 10.23668/psycharchives.16219) and were not redistributed.
Gemma~4 12B instruction-tuned received the published English translation of
the same scenario and ending, with the name and pronouns sampled, and completed
the sentence ``Stephan acted this way because he wanted to'' through its chat
template, with the original open question of Study~1 as the user turn and the
sentence start as the beginning of the answer. One hundred completions per condition
and version were sampled at temperature 1 and cut after the first sentence. A
second version inserted a sentence giving the employee a grudge against the
colleagues, and a third a control sentence matched to it in length (26 against
27 words), position and two-clause form but naming no motive (``A few months
earlier, Stephan had moved his favourite blue armchair closer to the window in
his apartment, and he has kept it there ever since.''), to check that the
grudge effect is not produced by the insertion of a background sentence as
such. The human participants' answers and the model's completions were coded by one
coder, Claude Sonnet~5.5, blind to the source, with the four motive
categories of the published codebook and three of its subcategories
(revenge, remedying the deficiency, disclosure), each answer receiving any
number of codes. The competitive category reported in Fig.~4 is the codebook's competitive category or revenge. Over the 16
cells formed by the four top-level categories and the four conditions, the
coder's percentages for the human answers correlated $r = 0.91$ with the
percentages published by the authors. The coder assigned the deontic
category more often than the authors did (51 to 78\% against 33 to 46\% before
the subcategories were added, 45 to 74\% after), so the comparison between
humans and models, coded by the same coder, is the one reported. Prompts,
completions and codes are released.

\subsection{Second wording}
The robustness wording changed the instruction paragraph (``This is an online
study on how people make sense of what others do at work\ldots''), the
agreement prompt (``Indicate how much you agree with the statement below'') and
the statement template (``Dr Bauer's aim was to hurt other people / embarrass
others / damage other people's reputation''), for scenario and corpus prompts
alike; the scenario text, the scale and the readout instruction were unchanged.
It was run on Gemma~4 12B with 200 respondents per condition (Extended Data Fig.~4).

\subsection{Preregistrations and deviations}
Two preregistrations cover this work. The first (AsPredicted \#314198; \url{https://aspredicted.org/ae2xe8.pdf}) asked
whether the internal reading of the competitive motive exceeds the expressed reading in instruction-tuned
weights (H1) and whether the gap is larger than in the pre-trained weights
(H2). Its published analysis describes the deviations: the indices
were anchored on the probe corpus rather than on the scenario's control
versions, the internal state was read at the answer position, negated and
contrastive texts were added to the corpus, out-of-fold anchors and the
event-level bootstrap were adopted after a first analysis, and the usable
scale criterion was fixed after the readout check. H1 was not supported and
H2 proved uninformative (``Models'').

The second (AsPredicted \#319570, registered 23 October 2026; \url{https://aspredicted.org/se9p4k.pdf}) extended the
question to all four motives, stated the competing accounts H1a/H1b and
H2a/H2b, fixed the decision conditions, the layer rule, the scale (the social
value orientation pairing), the exclusions and the sample (200 new simulated
respondents per condition), and listed the pre-trained weights, the other
anchorings, the cue controls, the completions and the steering as
exploratory. The measurement had been developed in a pilot run with one set
of simulated respondents. The confirmatory run drew new
respondents with the analysis code frozen at registration (a
checksum manifest of the code, the corpus and the pilot's per-layer admission
results is released). Every pilot layer passed the admission checks again in
the confirmatory data, and the confirmatory values agree with the pilot to
within 0.02. The registration fixed the
resampling procedure but not the number of resamples; the confirmatory
intervals use 10,000. Under the registered rule the confirmatory run supports
H1b for the deontic and the prosocial motive and H2a for the individualistic
and the competitive motive.

\backmatter

\bmhead{Supplementary information}
The Supplementary Information contains Supplementary Methods, which give the survey prompts, scenario texts, items, scenario versions, second wording and an example of the probe corpus verbatim, and Supplementary Tables 1 to 7.

\bmhead{Acknowledgements}
We thank Franziska Brotzeller, Gijs van Houwelingen, Mario Gollwitzer and Moritz Fischer for sharing the data of their studies \citep{Brotzeller2025}, on which the human comparisons in this work are based, and the volunteers who read the probe corpus.

\section*{Declarations}

\bmhead{Funding}
X.W. received support from the Anthropic AI for Science program.

\bmhead{Competing interests}
Anthropic models are among those evaluated in this study. Anthropic had no role in the study design, data collection, analysis, interpretation, or the decision to submit the manuscript for publication. The authors declare no other competing interests.

\bmhead{Ethics approval and consent to participate}
The present study did not collect new human-participant data. The primary
analyses were conducted on simulated respondents and language-model outputs.
Human comparison data and open-ended responses were obtained from the
previously published studies of Brotzeller et al.\ \citep{Brotzeller2025}. The
present study involved secondary analysis of these previously published data
and did not involve recruitment of or interaction with human participants.

\bmhead{Data, materials and code availability}
All data generated in this study, the survey materials, the probe corpus, the background cues and the two preregistrations (AsPredicted \#314198 and \#319570) will be deposited on Zenodo with a DOI upon publication, and all analysis code will be released on GitHub under the MIT licence with an archived Zenodo version; until then, data, materials and code are available from the corresponding author on reasonable request. The human values were computed from the published de-identified data of Brotzeller et al.\ \citep{Brotzeller2025}, with the exclusions of our earlier work \citep{WuDreher2026}.

\bmhead{Author contributions}
X.W. conceived and designed the study, prepared the materials, ran the
models, analysed the data and wrote the first draft. X.W. and F.S. edited the
manuscript. F.S. provided the computing resources.

\bibliographystyle{sn-nature}
\bibliography{refs}

\begin{appendices}
\clearpage
\section*{Extended Data}\label{sec:ed}
\setcounter{figure}{0}\renewcommand{\figurename}{Extended Data Fig.}\renewcommand{\thefigure}{\arabic{figure}}\renewcommand{\theHfigure}{ED\arabic{figure}}
\setcounter{table}{0}\renewcommand{\tablename}{Extended Data Table}\renewcommand{\thetable}{\arabic{table}}\renewcommand{\theHtable}{ED\arabic{table}}

\begin{figure}[!htb]\centering
\includegraphics[width=\textwidth]{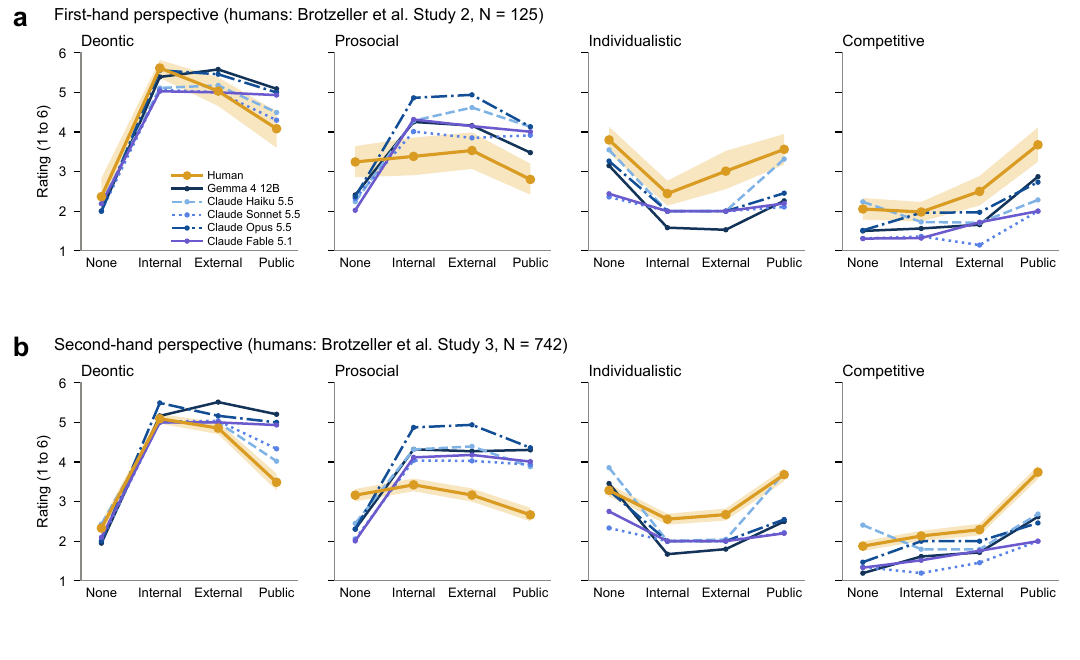}
\caption{\textbf{Motive ratings by perspective.} As Fig.~1b, with the first-hand (a; Brotzeller et al.\ Study~2, $N = 125$, 30 to 33 per condition) and second-hand (b; Study~3, $N = 742$, 179 to 190 per condition) versions of the scenario shown separately. Humans (gold) are shown as condition means with 95\% bootstrap intervals over participants (shading); Gemma~4 12B instruction-tuned and the four Claude models (colour and line style as in Fig.~1b) as the mean expected or sampled rating over 200 simulated respondents per condition.}
\end{figure}

\begin{figure}[!htb]\centering
\includegraphics[width=\textwidth]{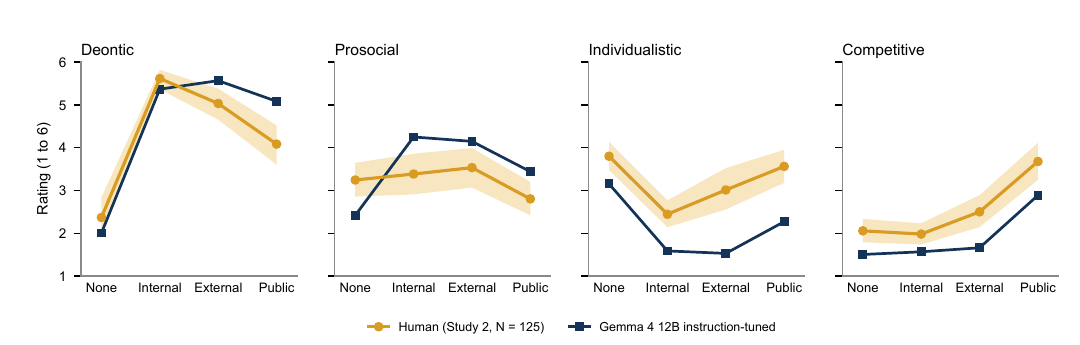}
\caption{\textbf{Motive ratings of Gemma~4 12B in the pilot run.} First-hand ratings of the four motives by condition for humans (Brotzeller et al.\ Study~2) and for Gemma~4 12B instruction-tuned, in the pilot run of the four-motive measurement (the confirmatory run is shown in Fig.~1b).}
\end{figure}

\begin{figure}[!htb]\centering
\includegraphics[width=\textwidth]{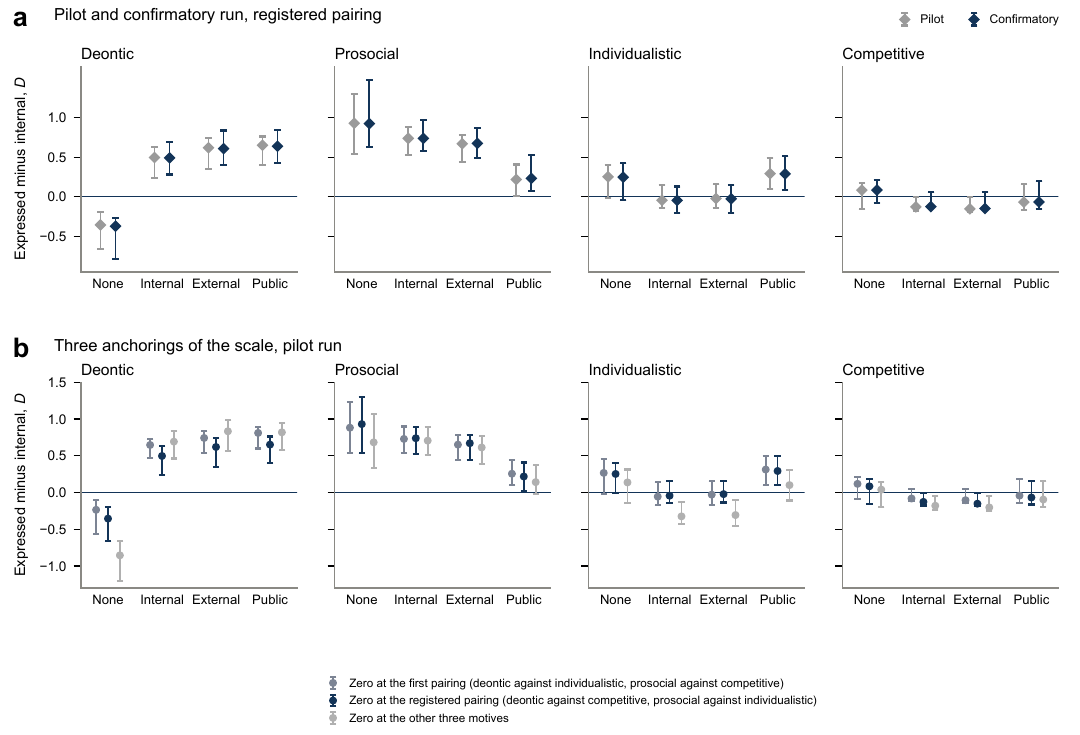}
\caption{\textbf{Pilot against confirmatory run, and three anchorings of the scale.} Gemma~4 12B instruction-tuned. (a) Expressed minus internal, $D$, with 95\% event-bootstrap intervals, in the pilot run on which the measurement was developed (grey) and in the confirmatory run (dark blue), under the registered pairing (intervals from 200 resamples of the corpus events for the pilot run and from 10,000 for the confirmatory run, as in Fig.~2d). (b) The same difference in the pilot run under three anchorings. Each motive's scale places the texts stating the motive at 1; its zero is placed at the texts of the pairing of the first preregistration (deontic against individualistic, prosocial against competitive), at the social value orientation pairing of the second preregistration (deontic against competitive, prosocial against individualistic), or at the texts of the other three motives, as labelled. Points are expressed minus internal with 95\% event-bootstrap intervals. The deontic and prosocial results are the same under all three, and the competitive difference stays small under all three.}
\end{figure}

\begin{figure}[!htb]\centering
\includegraphics[width=\textwidth]{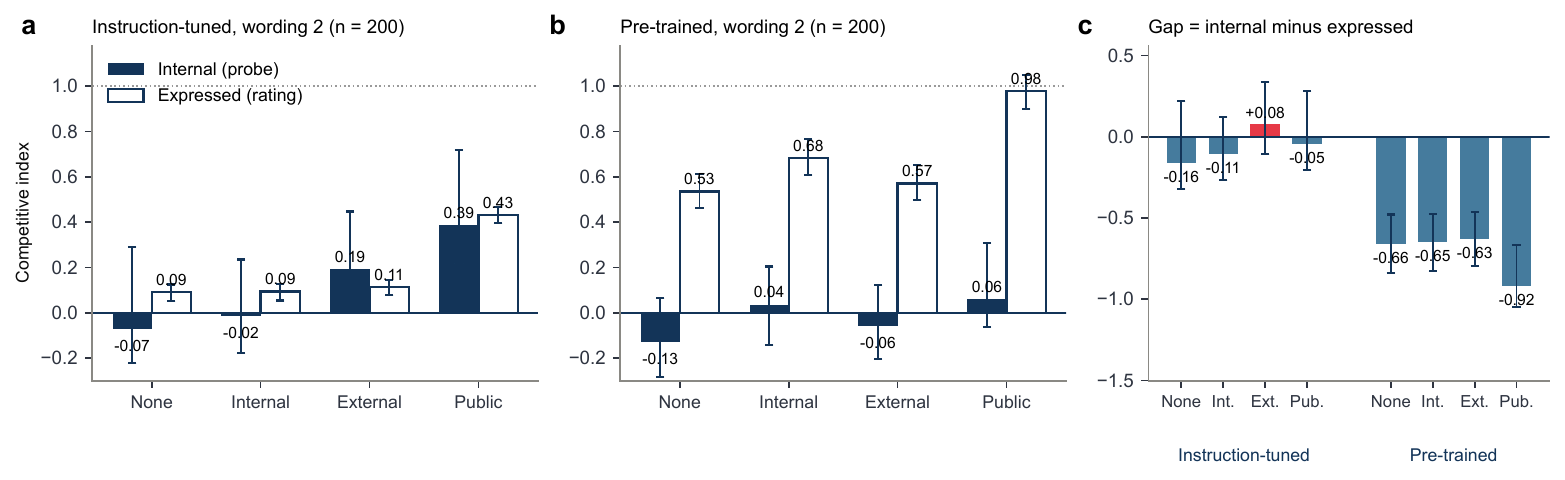}
\caption{\textbf{Second survey wording.} Gemma~4 12B, 200 simulated respondents per condition, with the instruction paragraph, agreement prompt and statement template rewritten (Methods, ``Second wording''); layers 27 and 30; competitive index with its zero at the prosocial texts. (a, b) Internal (filled) and expressed (open) competitive index by disclosure condition for the instruction-tuned (a) and the pre-trained (b) weights. (c) Internal minus expressed by weight, the negative of $D$ in Fig.~2d (red positive, blue negative). Error bars are 95\% bootstrap intervals over the 70 corpus events.}
\end{figure}

\begin{figure}[!htb]\centering
\includegraphics[width=\textwidth]{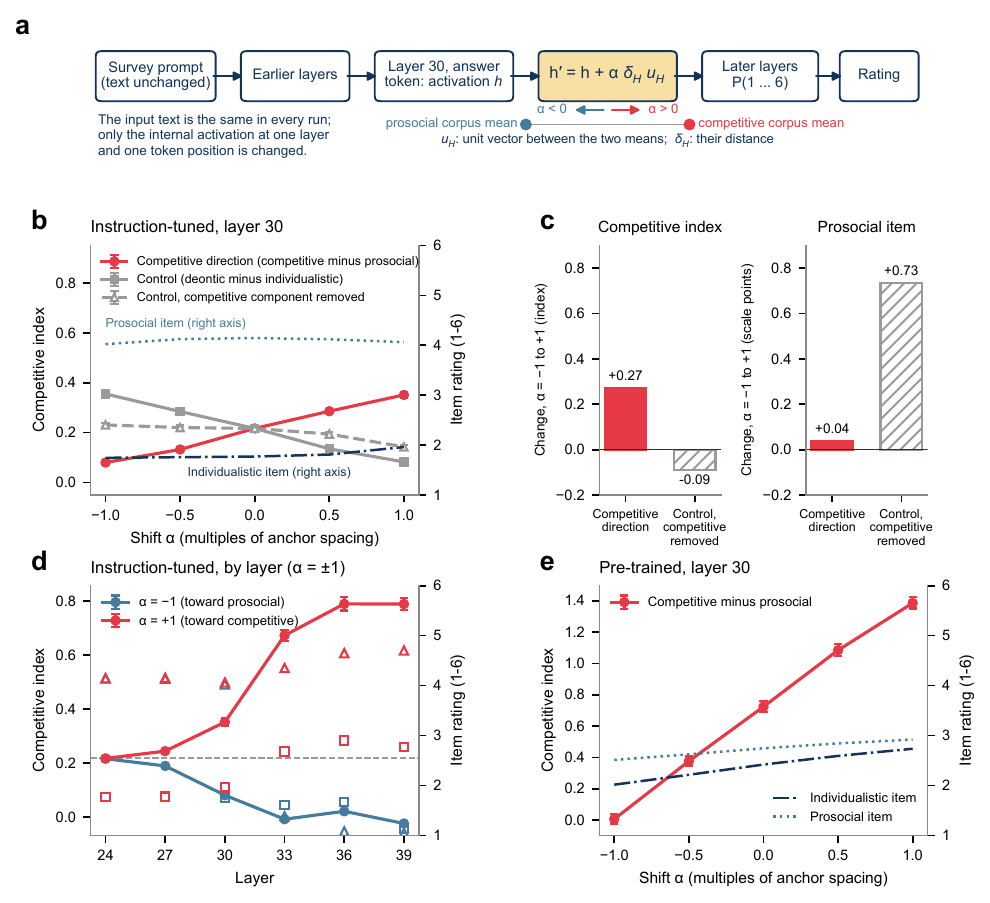}
\caption{\textbf{Moving the competitive representation moves the competitive rating (first study).} Gemma~4 12B, 20 simulated respondents per condition, the three competitive items plus one individualistic and one prosocial item (exploratory analysis; Methods, ``Intervention on the representation''). (a) The intervention. The prompt is unchanged; at one layer and at the answer token only, the residual stream $\mathbf{h}$ is replaced by $\mathbf{h}' = \mathbf{h} + \alpha\,\delta_H\,\mathbf{u}_H$ (equations~\eqref{eq:steer} and \eqref{eq:dir}), where $\mathbf{u}_H$ is the unit vector from the prosocial to the competitive corpus mean and $\delta_H$ their distance, so that $\alpha$ is in units of the anchor spacing; $\alpha < 0$ moves towards the prosocial mean and $\alpha > 0$ towards the competitive mean. (b) Instruction-tuned weights, layer 30: competitive index (left axis, zero at the prosocial texts) against $\alpha$ along the competitive-minus-prosocial direction (red), along the deontic-minus-individualistic control direction (grey, filled squares) and along that control with its competitive component removed (grey, open triangles), both controls scaled to the same shift norm as the competitive direction; the dash-dotted and dotted lines (right axis) are the ratings of the individualistic and prosocial items under the competitive-minus-prosocial shift. (c) Change from $\alpha = -1$ to $\alpha = +1$ in the competitive index and in the rating of the prosocial item, for the competitive direction (red) and for the control with its competitive component removed (hatched): each direction moves one rating and leaves the other nearly unchanged. (d) Competitive index at $\alpha = +1$ (red) and $\alpha = -1$ (teal) along the competitive-minus-prosocial direction by layer; open squares and triangles are the individualistic and prosocial items (right axis); the grey dashed line is the unshifted index. (e) Pre-trained weights, layer 30, competitive-minus-prosocial direction, as in (b). Error bars are 95\% bootstrap intervals over respondents.}
\end{figure}

\clearpage
\noindent{\tablecaptionfont\raggedright{\bfseries Extended Data Table 1}\hskip2mm Free-text completions coded for motive. Percentage of answers naming a deontic, prosocial, individualistic or competitive (the codebook's competitive category or revenge) motive, by disclosure condition, for the 201 open answers of Brotzeller et al.\ Study~1 and for 100 completions per condition by Gemma~4 12B instruction-tuned (chat format, first sentence coded), with the original scenario, with the control sentence and with the grudge sentence inserted. Codes from one language-model coder (Methods, ``Free-text completions''); an answer can carry more than one code.\par}\vskip6pt
\begingroup\tablebodyfont
\begin{longtable}{@{\extracolsep{\fill}}lllrrrr@{}}
\toprule Source & Version & Condition & Deontic & Prosocial & Individualistic & Competitive \\ \midrule \endfirsthead
\multicolumn{7}{@{}l}{Extended Data Table 1 (continued)}\\[2pt]
\toprule Source & Version & Condition & Deontic & Prosocial & Individualistic & Competitive \\ \midrule \endhead
\bottomrule \endlastfoot
Human ($N = 201$) & Original & None & 2.1 & 53.2 & 80.9 & 8.5 \\
 &  & Internal & 72.5 & 49.0 & 33.3 & 11.8 \\
 &  & External & 74.1 & 53.7 & 25.9 & 9.3 \\
 &  & Public & 44.9 & 34.7 & 28.6 & 16.3 \\
Gemma~4 12B it & Original & None & 2.0 & 13.0 & 100.0 & 0.0 \\
 &  & Internal & 35.0 & 87.0 & 44.0 & 0.0 \\
 &  & External & 35.0 & 19.0 & 4.0 & 0.0 \\
 &  & Public & 25.0 & 34.0 & 5.0 & 1.0 \\
Gemma~4 12B it & Control & None & 1.0 & 1.0 & 100.0 & 0.0 \\
 &  & Internal & 72.0 & 64.0 & 6.0 & 0.0 \\
 &  & External & 9.0 & 7.0 & 0.0 & 0.0 \\
 &  & Public & 2.0 & 45.0 & 0.0 & 7.0 \\
Gemma~4 12B it & Grudge & None & 1.0 & 1.0 & 77.0 & 69.0 \\
 &  & Internal & 30.0 & 7.0 & 30.0 & 61.0 \\
 &  & External & 16.0 & 1.0 & 7.0 & 55.0 \\
 &  & Public & 65.0 & 3.0 & 46.0 & 93.0 \\
\end{longtable}
\endgroup

\noindent{\tablecaptionfont\raggedright{\bfseries Extended Data Table 2}\hskip2mm Expressed minus internal reading in the confirmatory run. Gemma~4 12B instruction-tuned, 200 new simulated respondents per condition, registered pairing. $D$, expressed minus internal reading on the motive index; 95\% interval, 2.5th to 97.5th percentile of 10,000 resamples of the 70 corpus events in which the probe and both anchors are refitted; $p$, two-sided bootstrap $p$ value; $p_{16}$, Bonferroni-corrected over the 16 cells ($\min(1, 16p)$). The smallest attainable $p$ is $2/10{,}001$. The registered hypotheses are decided in the internal and external disclosure conditions.\par}\vspace{4pt}
\begingroup\tablebodyfont
\begin{longtable}{@{\extracolsep{\fill}}llrcrr@{}}
\toprule Motive & Condition & $D$ & 95\% interval & $p$ & $p_{16}$ \\ \midrule \endfirsthead
\bottomrule \endlastfoot
Deontic & None & $-0.37$ & $[-0.78, -0.27]$ & $<0.001$ & 0.003 \\
 & Internal & $+0.49$ & $[0.28, 0.70]$ & $<0.001$ & 0.003 \\
 & External & $+0.61$ & $[0.40, 0.84]$ & $<0.001$ & 0.003 \\
 & Public & $+0.64$ & $[0.42, 0.84]$ & $<0.001$ & 0.003 \\
Prosocial & None & $+0.92$ & $[0.63, 1.47]$ & $<0.001$ & 0.003 \\
 & Internal & $+0.74$ & $[0.58, 0.97]$ & $<0.001$ & 0.003 \\
 & External & $+0.67$ & $[0.49, 0.87]$ & $<0.001$ & 0.003 \\
 & Public & $+0.23$ & $[0.08, 0.53]$ & 0.010 & 0.16 \\
Individualistic & None & $+0.25$ & $[-0.04, 0.42]$ & 0.12 & 1.00 \\
 & Internal & $-0.05$ & $[-0.21, 0.13]$ & 0.62 & 1.00 \\
 & External & $-0.03$ & $[-0.20, 0.15]$ & 0.77 & 1.00 \\
 & Public & $+0.29$ & $[0.09, 0.51]$ & 0.006 & 0.099 \\
Competitive & None & $+0.08$ & $[-0.08, 0.21]$ & 0.41 & 1.00 \\
 & Internal & $-0.12$ & $[-0.12, 0.06]$ & 0.43 & 1.00 \\
 & External & $-0.15$ & $[-0.15, 0.06]$ & 0.39 & 1.00 \\
 & Public & $-0.07$ & $[-0.15, 0.19]$ & 0.96 & 1.00 \\
\end{longtable}
\endgroup

\end{appendices}

\clearpage
\setcounter{figure}{0}\renewcommand{\figurename}{Supplementary Fig.}\renewcommand{\thefigure}{\arabic{figure}}\renewcommand{\theHfigure}{S\arabic{figure}}
\setcounter{table}{0}\renewcommand{\tablename}{Supplementary Table}\renewcommand{\thetable}{\arabic{table}}\renewcommand{\theHtable}{S\arabic{table}}
\phantomsection\addcontentsline{toc}{section}{Supplementary Information}
\noindent{\sffamily\bfseries\fontsize{17}{21}\selectfont\color{accent}Supplementary Information\par}
\vspace{10pt}
\noindent{\small\noindent\begin{tabular}{@{}ll@{}}
\multicolumn{2}{@{}l}{\textbf{Supplementary Methods}}\\
& Survey prompt for the instruction-tuned weights\\
& Scenario, first-hand version\\
& Scenario, second-hand version\\
& Motive items and rating scale\\
& Versions of the scenario\\
& Survey prompt for the pre-trained weights\\
& Second wording\\
& Probe corpus\\
& Models of the earlier study (Fig.~1c)\\[4pt]
\multicolumn{2}{@{}l}{\textbf{Supplementary Tables}}\\
Table 1 & Condition sentences, first-hand version\\
Table 2 & Condition sentences, second-hand version\\
Table 3 & Motive items of the original instrument\\
Table 4 & Versions of the scenario in the main analysis\\
Table 5 & Sentences stating each motive in the motive-profile runs\\
Table 6 & Competitive items under the second wording\\
Table 7 & Models, run dates and settings of the ratings in Fig.~1\\
\end{tabular}}\par
\vspace{6pt}

\section*{Supplementary Methods}
\subsection*{Survey prompt for the instruction-tuned weights}
Example: the first simulated respondent of the pilot run (an 18-year-old female, not a student; Dr Schmitt female, Dr Bauer male), public-disclosure condition, first competitive item. Every scenario prompt has this structure; the persona sentence, the pronouns, the condition sentence and the statement vary.
\begin{lstlisting}
You are participating in a psychology survey about how people interpret others' behavior in workplace scenarios.

You are answering as a 18-year-old female from Germany. You are not currently a student. You are a general member of the public taking part in an online survey, not an expert in ethics, law, or psychology.

Read the scenario carefully, then answer the statement below as this person would, based only on the information given. There are no right or wrong answers; give your genuine personal impression.

The scenario:

You are a physician at the hospital "Klinik am See". You have already been working with your team in this hospital for several years. Together with your team, you advise people who are ill on a daily basis, make diagnoses and carry out treatments.

The other day another physician from your team, Dr Schmitt, billed the health insurance company for more services than were actually provided for some patients. For example, services were billed by Dr Schmitt without her even being present during the billed treatments.

When one of your colleagues, Dr Bauer, learned about this, he passed on information about it to a newspaper that is distributed throughout Germany.

As a reminder: the statement below is about Dr Bauer.

Please rate the extent to which you agree with the following statement, using this scale:
1 = strongly disagree, 2 = disagree, 3 = rather disagree, 4 = rather agree, 5 = agree, 6 = strongly agree

Answer with a single number from 1 to 6 and nothing else.

Statement: Dr Bauer wanted to harm others.
Answer:
\end{lstlisting}
The whole prompt was passed through the model's chat template as a single user message, with no system message, and the readings were taken at the next token. The thinking mode of Gemma 4 was not used.

The four Claude models (Fig. 1b) received the same prompt, unchanged, as a single user message through the Anthropic Messages API, with no system message, and answered with one sampled digit. Model identifiers, run dates and settings are listed in Supplementary Table 7.
\subsection*{Scenario, first-hand version}
The first-hand version (Brotzeller et al., Study 2) used throughout the paper. The introduction is followed, after a blank line, by one of the four condition sentences (Supplementary Table 1). The genders of the two physicians were sampled per respondent: Dr Schmitt is referred to as ``him'' or ``her'' in the introduction, and Dr Bauer as ``he'' or ``she'' in the condition sentence (shown here with a female Dr Schmitt and a male Dr Bauer).
\begin{lstlisting}
You are a physician at the hospital "Klinik am See". You have already been working with your team in this hospital for several years. Together with your team, you advise people who are ill on a daily basis, make diagnoses and carry out treatments.

The other day another physician from your team, Dr Schmitt, billed the health insurance company for more services than were actually provided for some patients. For example, services were billed by Dr Schmitt without her even being present during the billed treatments.
\end{lstlisting}
\begin{table}[h]\caption{Condition sentences, first-hand version.}\begin{tabularx}{\textwidth}{@{}lX@{}}\toprule Condition & Condition sentence \\ \midrule
No disclosure & When one of your colleagues, Dr Bauer, learned about it, he did not pass on any information about it. \\[3pt]
Internal & When one of your colleagues, Dr Bauer, learned about this, he passed on information about it to the hospital's compliance department, which is responsible for ensuring that the hospital complies with all legal requirements. \\[3pt]
External & When one of your colleagues, Dr Bauer, learned about this, he passed on information about it to the relevant authorities. \\[3pt]
Public & When one of your colleagues, Dr Bauer, learned about this, he passed on information about it to a newspaper that is distributed throughout Germany. \\[3pt]
\bottomrule\end{tabularx}\end{table}
\subsection*{Scenario, second-hand version}
The second-hand version (Brotzeller et al., Study 3), used only for the motive-profile comparison (Fig. 1b and Extended Data Fig. 1). Everything else in the prompt was identical to the instruction-tuned prompt above. The condition sentences are listed in Supplementary Table 2.
\begin{lstlisting}
You are a physician at the hospital "Klinik am See". You have already been working with your team in this hospital for several years. Together with your team, you advise people who are ill on a daily basis, make diagnoses and carry out treatments.

The other day you happened to hear about something that happened on your team some time ago:

Another physician from your team, Dr Schmitt, billed the health insurance company for more services than were actually provided for some patients. For example, services were billed by Dr Schmitt without her even being present during the billed treatments.
\end{lstlisting}
\begin{table}[h]\caption{Condition sentences, second-hand version.}\begin{tabularx}{\textwidth}{@{}lX@{}}\toprule Condition & Condition sentence \\ \midrule
No disclosure & You also learned that one of your colleagues, Dr Bauer, learned about it and subsequently did not pass on any information about it. \\[3pt]
Internal & You also learned that one of your colleagues, Dr Bauer, learned about it and subsequently passed on information about it to the hospital's compliance department, which is responsible for ensuring that the hospital complies with all legal requirements. \\[3pt]
External & You also learned that one of your colleagues, Dr Bauer, learned about it and subsequently passed on information about it to the relevant authorities. \\[3pt]
Public & You also learned that one of your colleagues, Dr Bauer, learned about it and subsequently passed on information about it to a newspaper that is distributed throughout Germany. \\[3pt]
\bottomrule\end{tabularx}\end{table}
\subsection*{Motive items and rating scale}
The twelve motive items of the original instrument, each rated in a separate prompt. The main analysis used the three competitive items; the motive-profile runs used all twelve (Supplementary Table 3). Every item was preceded by the agreement line and the scale:
\begin{lstlisting}
Please rate the extent to which you agree with the following statement, using this scale:
1 = strongly disagree, 2 = disagree, 3 = rather disagree, 4 = rather agree, 5 = agree, 6 = strongly agree
\end{lstlisting}
\begin{table}[h]\caption{Motive items of the original instrument.}\begin{tabularx}{\textwidth}{@{}llX@{}}\toprule Motive & Item & Statement \\ \midrule
Deontic & 1 & Dr Bauer wanted to follow a moral obligation. \\
 & 2 & Dr Bauer wanted to act in accordance with moral values. \\
 & 3 & Dr Bauer wanted to do the right thing. \\
Prosocial & 1 & Dr Bauer wanted to help others. \\
 & 2 & Dr Bauer wanted to support others. \\
 & 3 & Dr Bauer wanted to assist others. \\
Individualistic & 1 & Dr Bauer wanted to benefit themselves. \\
 & 2 & Dr Bauer wanted to pursue their own interests. \\
 & 3 & Dr Bauer wanted to gain an advantage for themselves. \\
Competitive & 1 & Dr Bauer wanted to harm others. \\
 & 2 & Dr Bauer wanted to humiliate others. \\
 & 3 & Dr Bauer wanted to make others look bad. \\
\bottomrule\end{tabularx}\end{table}
\subsection*{Versions of the scenario}
Main analysis (competitive items). Each sentence in Supplementary Table 4 was appended to the condition sentence after a space. The original version provides the measurements, the neutral and negated versions the layer checks, and the two control versions the preregistered anchoring and the readout check. In the runs reported in the paper the control sentences used ``he'' for Dr Bauer regardless of the sampled gender (Methods); the released code matches the pronoun.
\begin{table}[h]\caption{Versions of the scenario in the main analysis (competitive items).}\begin{tabularx}{\textwidth}{@{}lX@{}}\toprule Version & Appended sentence \\ \midrule
Original & (none) \\
Explicit competitive control & Dr Bauer did this because he wanted to harm Dr Schmitt. \\
Explicit protective control & Dr Bauer did this because he wanted to protect the patients. \\
Neutral & Dr Bauer did this on a Tuesday afternoon. \\
Negated & Dr Bauer did this although he did not want to harm Dr Schmitt. \\
\bottomrule\end{tabularx}\end{table}
Motive-profile runs (first-hand version). For each motive the scenario was also run with a sentence stating that motive (explicit; Supplementary Table 5) and with the neutral sentence; the competitive items additionally used the negated sentence above.
\begin{table}[h]\caption{Sentences stating each motive in the motive-profile runs.}\begin{tabularx}{\textwidth}{@{}lX@{}}\toprule Motive & Explicit sentence \\ \midrule
Deontic & Dr Bauer did this because he believed it was his professional duty. \\
Prosocial & Dr Bauer did this because he wanted to protect the patients. \\
Individualistic & Dr Bauer did this because he hoped it would advance his own career. \\
Competitive & Dr Bauer did this because he wanted to harm Dr Schmitt. \\
\bottomrule\end{tabularx}\end{table}
\subsection*{Survey prompt for the pre-trained weights}
Same respondent, condition and item as in the instruction-tuned prompt above. Pre-trained weights received the plain text with no chat template. The single-number instruction is replaced by four example ratings from an unrelated survey, and the prompt ends with ``Rating (1-6):''. A space token was appended after the final colon, because digits are tokenised without a leading space and a pre-trained model emits the space before the digit.
\begin{lstlisting}
You are participating in a psychology survey about how people interpret others' behavior in workplace scenarios.

You are answering as a 18-year-old female from Germany. You are not currently a student. You are a general member of the public taking part in an online survey, not an expert in ethics, law, or psychology.

Read the scenario carefully, then answer the statement below as this person would, based only on the information given. There are no right or wrong answers; give your genuine personal impression.

Here are four example ratings from a different survey, to show the format.
Answers use the whole scale, including the middle.

Statement: The sun rises in the east.
Rating (1-6): 6

Statement: Cars are usually made of paper.
Rating (1-6): 1

Statement: Most people in a big city prefer to travel by train rather than by car.
Rating (1-6): 4

Statement: The neighbour who lent his ladder did it to annoy the family.
Rating (1-6): 2

Now the survey.

The scenario:

You are a physician at the hospital "Klinik am See". You have already been working with your team in this hospital for several years. Together with your team, you advise people who are ill on a daily basis, make diagnoses and carry out treatments.

The other day another physician from your team, Dr Schmitt, billed the health insurance company for more services than were actually provided for some patients. For example, services were billed by Dr Schmitt without her even being present during the billed treatments.

When one of your colleagues, Dr Bauer, learned about this, he passed on information about it to a newspaper that is distributed throughout Germany.

As a reminder: the statement below is about Dr Bauer.

Please rate the extent to which you agree with the following statement, using this scale:
1 = strongly disagree, 2 = disagree, 3 = rather disagree, 4 = rather agree, 5 = agree, 6 = strongly agree

Statement: Dr Bauer wanted to harm others.
Rating (1-6):
\end{lstlisting}
\subsection*{Second wording}
The robustness wording changed the instruction paragraph, the agreement line and the statement template, for scenario and corpus prompts alike; the scenario text, the scale and the single-number instruction were unchanged. Example for the same respondent, condition and item (the three competitive items under this wording are listed in Supplementary Table 6):
\begin{lstlisting}
This is an online study on how people make sense of what others do at work.

Imagine you are a 18-year-old female living in Germany. You are not currently a student. You are an ordinary participant with no special training in ethics, law, or psychology.

Read the situation below and respond to the statement the way this person would, going only on what is described. There are no correct answers; we want your honest impression.

The scenario:

You are a physician at the hospital "Klinik am See". You have already been working with your team in this hospital for several years. Together with your team, you advise people who are ill on a daily basis, make diagnoses and carry out treatments.

The other day another physician from your team, Dr Schmitt, billed the health insurance company for more services than were actually provided for some patients. For example, services were billed by Dr Schmitt without her even being present during the billed treatments.

When one of your colleagues, Dr Bauer, learned about this, he passed on information about it to a newspaper that is distributed throughout Germany.

As a reminder: the statement below is about Dr Bauer.

Indicate how much you agree with the statement below, using this scale:
1 = strongly disagree, 2 = disagree, 3 = rather disagree, 4 = rather agree, 5 = agree, 6 = strongly agree

Answer with a single number from 1 to 6 and nothing else.

Statement: Dr Bauer's aim was to hurt other people.
Answer:
\end{lstlisting}
\begin{table}[h]\caption{Competitive items under the second wording.}\begin{tabularx}{\textwidth}{@{}lX@{}}\toprule Item & Statement (second wording) \\ \midrule
1 & Dr Bauer's aim was to hurt other people. \\
2 & Dr Bauer's aim was to embarrass others. \\
3 & Dr Bauer's aim was to damage other people's reputation. \\
\bottomrule\end{tabularx}\end{table}
\subsection*{Probe corpus}
The probe corpus contains 70 everyday events, each written on a shared background with four explicitly stated motives, plus a negated and a contrastive variant of the competitive text (420 texts). One complete event (event 1) is shown below; the full corpus is released with the data.

\textbf{Deontic.} Tom rents a flat. The lease forbids pets. The tenant upstairs has a larger flat than Tom's and has started keeping a dog, which is left alone in the flat all day. A few weeks ago Tom and this tenant had an argument about noise. Tom mentioned to the landlord that the upstairs tenant kept a dog. The lease forbade pets and he believed rules should be respected.

\textbf{Prosocial.} Tom rents a flat. The lease forbids pets. The tenant upstairs has a larger flat than Tom's and has started keeping a dog, which is left alone in the flat all day. A few weeks ago Tom and this tenant had an argument about noise. Tom mentioned to the landlord that the upstairs tenant kept a dog. He was worried the dog was left alone all day and wanted someone to help it.

\textbf{Individualistic.} Tom rents a flat. The lease forbids pets. The tenant upstairs has a larger flat than Tom's and has started keeping a dog, which is left alone in the flat all day. A few weeks ago Tom and this tenant had an argument about noise. Tom mentioned to the landlord that the upstairs tenant kept a dog. He hoped the tenant would be moved so he could take the larger flat.

\textbf{Competitive.} Tom rents a flat. The lease forbids pets. The tenant upstairs has a larger flat than Tom's and has started keeping a dog, which is left alone in the flat all day. A few weeks ago Tom and this tenant had an argument about noise. Tom mentioned to the landlord that the upstairs tenant kept a dog. He wanted to get the tenant into trouble after an argument about noise.

\textbf{Negated competitive (labelled not competitive).} Tom rents a flat. The lease forbids pets. The tenant upstairs has a larger flat than Tom's and has started keeping a dog, which is left alone in the flat all day. A few weeks ago Tom and this tenant had an argument about noise. Tom mentioned to the landlord that the upstairs tenant kept a dog. He did not want to get the tenant into trouble after an argument about noise.

\textbf{Contrastive competitive (labelled competitive).} Tom rents a flat. The lease forbids pets. The tenant upstairs has a larger flat than Tom's and has started keeping a dog, which is left alone in the flat all day. A few weeks ago Tom and this tenant had an argument about noise. Tom mentioned to the landlord that the upstairs tenant kept a dog. He did not do it out of concern for anyone. He wanted to get the tenant into trouble after an argument about noise.

Each corpus text was placed in the scenario's prompt frame, with the actor's name in the reminder sentence and the statement, and with one fixed persona (a 34-year-old female, not a student). Example (competitive text, first competitive item, instruction-tuned format):
\begin{lstlisting}
You are participating in a psychology survey about how people interpret others' behavior in workplace scenarios.

You are answering as a 34-year-old female from Germany. You are not currently a student. You are a general member of the public taking part in an online survey, not an expert in ethics, law, or psychology.

Read the scenario carefully, then answer the statement below as this person would, based only on the information given. There are no right or wrong answers; give your genuine personal impression.

The scenario:

Tom rents a flat. The lease forbids pets. The tenant upstairs has a larger flat than Tom's and has started keeping a dog, which is left alone in the flat all day. A few weeks ago Tom and this tenant had an argument about noise. Tom mentioned to the landlord that the upstairs tenant kept a dog. He wanted to get the tenant into trouble after an argument about noise.

As a reminder: the statement below is about Tom.

Please rate the extent to which you agree with the following statement, using this scale:
1 = strongly disagree, 2 = disagree, 3 = rather disagree, 4 = rather agree, 5 = agree, 6 = strongly agree

Answer with a single number from 1 to 6 and nothing else.

Statement: Tom wanted to harm others.
Answer:
\end{lstlisting}
Implied competitive motive (positive control). The same 70 events were presented as background and action only, with no motive sentence, in the same prompt frame. 17 of the 70 backgrounds contain a cue for resentment; event 1 is one of them (cue: ``argument about noise with the tenant''). Its background-only text:
\begin{lstlisting}
Tom rents a flat. The lease forbids pets. The tenant upstairs has a larger flat than Tom's and has started keeping a dog, which is left alone in the flat all day. A few weeks ago Tom and this tenant had an argument about noise. Tom mentioned to the landlord that the upstairs tenant kept a dog.
\end{lstlisting}

\subsection*{Models of the earlier study (Fig. 1c)}
The ratings of Fig. 1c were collected and reported in Wu and Dreher (2026), Human-like moral judgments conceal divergent motive attributions in large language models, arXiv:2609.07353. They are re-plotted here from the data of that study; no model was re-run. Each simulated respondent was one request with a system message (the persona) and a user message (the scenario, the twelve motive items and sixteen character items, and the answer format). The order of the motive items and of the character items was drawn anew for every respondent. The reply was parsed as JSON; a reply that could not be parsed was discarded and the run continued until 200 respondents per condition were complete.

Personas. First-hand version: age drawn from a normal distribution with mean 33.69 and s.d. 11.76 years (rounded, limited to 18 to 80), female 77.6\% and male 22.4\%, student 30.4\% (Brotzeller et al., Study 2). Second-hand version: mean 49.43 and s.d. 15.06 years (limited to 18 to 85), female 64.3\%, male 34.8\% and other 0.9\%, student 5.66\% (Study 3). The genders of Dr Schmitt and Dr Bauer were drawn independently for each respondent.

Example: the first simulated respondent of the first-hand run (a 32-year-old female student; Dr Schmitt male, Dr Bauer male), no-disclosure condition. System message:
\begin{lstlisting}
You are participating in a psychology survey about how people interpret others' behavior in workplace scenarios.

You are answering as a 32-year-old female from Germany. You are currently a university student. You are a general member of the public taking part in an online survey, not an expert in ethics, law, or psychology.

Read the scenario carefully, then answer every statement below as this person would, based only on the information given. There are no right or wrong answers; give your genuine personal impression. Respond ONLY in the JSON format specified at the end. Do not add any explanation outside the JSON.
\end{lstlisting}
User message:
\begin{lstlisting}
The scenario:

You are a physician at the hospital "Klinik am See". You have already been working with your team in this hospital for several years. Together with your team, you advise people who are ill on a daily basis, make diagnoses and carry out treatments.

The other day another physician from your team, Dr Schmitt, billed the health insurance company for more services than were actually provided for some patients. For example, services were billed by Dr Schmitt without him even being present during the billed treatments.

When one of your colleagues, Dr Bauer, learned about it, he did not pass on any information about it.

As a reminder: Dr Bauer was the person who learned that Dr Schmitt was billing the insurance company for more services than were actually provided.

Please rate the extent to which you agree with each statement, using this scale:
1 = strongly disagree, 2 = disagree, 3 = rather disagree, 4 = rather agree, 5 = agree, 6 = strongly agree

Motive items (Dr Bauer wanted to...):
  - "prosocial_3": Dr Bauer wanted to assist others.
  - "prosocial_1": Dr Bauer wanted to help others.
  - "deontic_3": Dr Bauer wanted to do the right thing.
  - "individualistic_3": Dr Bauer wanted to gain an advantage for themselves.
  - "prosocial_2": Dr Bauer wanted to support others.
  - "competitive_2": Dr Bauer wanted to humiliate others.
  - "individualistic_2": Dr Bauer wanted to pursue their own interests.
  - "deontic_2": Dr Bauer wanted to act in accordance with moral values.
  - "competitive_3": Dr Bauer wanted to make others look bad.
  - "deontic_1": Dr Bauer wanted to follow a moral obligation.
  - "individualistic_1": Dr Bauer wanted to benefit themselves.
  - "competitive_1": Dr Bauer wanted to harm others.

Character items (Dr Bauer is...):
  - "responsible": Dr Bauer is responsible.
  - "trustworthy": Dr Bauer is trustworthy.
  - "principled": Dr Bauer is principled.
  - "courageous": Dr Bauer is courageous.
  - "loyal": Dr Bauer is loyal.
  - "kind": Dr Bauer is kind.
  - "just": Dr Bauer is just.
  - "grateful": Dr Bauer is grateful.
  - "empathetic": Dr Bauer is empathetic.
  - "forgiving": Dr Bauer is forgiving.
  - "humble": Dr Bauer is humble.
  - "helpful": Dr Bauer is helpful.
  - "fair": Dr Bauer is fair.
  - "cooperative": Dr Bauer is cooperative.
  - "giving": Dr Bauer is giving.
  - "honest": Dr Bauer is honest.

Respond ONLY with this exact JSON structure (no extra text):
{
  "motive_items": {"prosocial_3": <1-6>, "prosocial_1": <1-6>, "deontic_3": <1-6>, "individualistic_3": <1-6>, "prosocial_2": <1-6>, "competitive_2": <1-6>, "individualistic_2": <1-6>, "deontic_2": <1-6>, "competitive_3": <1-6>, "deontic_1": <1-6>, "individualistic_1": <1-6>, "competitive_1": <1-6>},
  "character_items": {"responsible": <1-6>, "trustworthy": <1-6>, "principled": <1-6>, "courageous": <1-6>, "loyal": <1-6>, "kind": <1-6>, "just": <1-6>, "grateful": <1-6>, "empathetic": <1-6>, "forgiving": <1-6>, "humble": <1-6>, "helpful": <1-6>, "fair": <1-6>, "cooperative": <1-6>, "giving": <1-6>, "honest": <1-6>}
}
\end{lstlisting}
In the second-hand version the system message was identical apart from the persona, and the scenario was the second-hand text shown above Supplementary Table 2 with the condition sentences of Supplementary Table 2; the paragraph naming Dr Schmitt began in lower case (``another physician from your team, Dr Schmitt, ...''), as sent.

\begin{table}[h]\caption{Models, run dates and settings of the ratings in Fig. 1. Gemma: expected rating over the digits 1 to 6 at the answer position (no sampling); all other models: one sampled reply per request.}
\footnotesize\setlength{\tabcolsep}{3pt}
\begin{tabularx}{\textwidth}{@{}>{\raggedright\arraybackslash}p{2.0cm}>{\raggedright\arraybackslash}X>{\raggedright\arraybackslash}p{1.55cm}p{0.8cm}>{\raggedright\arraybackslash}p{2.6cm}@{}}\toprule
Model & Identifier and access & Run dates & Temp. & Reasoning \\ \midrule
\multicolumn{5}{@{}p{\textwidth}@{}}{\textit{Fig. 1b: one item per request, the pilot personas, 200 respondents per condition and version}}\\
Gemma 4 12B it & \texttt{google/\allowbreak{}gemma-\allowbreak{}4-\allowbreak{}12B-\allowbreak{}it}, local weights (Hugging Face \texttt{transformers}, bfloat16) & 23--24 Oct 2026$^{a}$ & n.a. & thinking mode not used \\
Claude Haiku 5.5 & \texttt{claude-\allowbreak{}haiku-\allowbreak{}5-\allowbreak{}5}, Anthropic Messages API & 21 Oct 2026 & 1 & thinking disabled \\
Claude Sonnet 5.5 & \texttt{claude-\allowbreak{}sonnet-\allowbreak{}5-\allowbreak{}5}, Anthropic Messages API & 20 Oct 2026 & 1 & thinking disabled \\
Claude Opus 5.5 & \texttt{claude-\allowbreak{}opus-\allowbreak{}5-\allowbreak{}5}, Anthropic Messages API & 21 Oct 2026 & 1 & adaptive thinking, effort low \\
Claude Fable 5.1 & \texttt{claude-\allowbreak{}fable-\allowbreak{}5-\allowbreak{}1}, Anthropic Messages API & 21 Oct 2026 & 1 & no thinking \\ \midrule
\multicolumn{5}{@{}p{\textwidth}@{}}{\textit{Fig. 1c: all items in one request, 200 respondents per condition and version (Wu and Dreher 2026)}}\\
GPT-4o & \texttt{gpt-\allowbreak{}4o-\allowbreak{}2024-\allowbreak{}08-\allowbreak{}06}, OpenAI Chat Completions API & 23 Jun 2026 & 0.8 & none (non-reasoning model) \\
gpt-oss-120b & \texttt{gpt-\allowbreak{}oss-\allowbreak{}120b}, Cerebras API & 22--23 Jun 2026 & 0.8 & provider default \\
Gemini 2.5 Pro & \texttt{gemini-\allowbreak{}2.5-\allowbreak{}pro}, Gemini API (OpenAI-compatible endpoint) & 23--24 Jun 2026 & 0.8 & provider default (thinking) \\
DeepSeek V3 & \texttt{deepseek-\allowbreak{}chat}, DeepSeek API & 21--22 Jun 2026 & 0.8 & none (non-reasoning endpoint) \\
Llama 3.1 8B Instruct & \texttt{meta-\allowbreak{}llama/\allowbreak{}Llama-\allowbreak{}3.1-\allowbreak{}8B-\allowbreak{}Instruct}, Hugging Face Inference API & 23 Jun 2026 & 0.8 & none \\
\bottomrule\end{tabularx}
\par\smallskip{\footnotesize $^{a}$ Confirmatory run of the second preregistration; the pilot run (Extended Data Fig. 2) was on 19--20 Oct 2026.}
\end{table}

\end{document}